%% file: main.tex
\documentclass[journal]{IEEEtran}
\usepackage{subcaption}
\usepackage{amsmath,amsfonts}
\usepackage{algorithmic}
\usepackage{array}
\usepackage{textcomp}
\usepackage{stfloats}
\usepackage{url}
\usepackage{verbatim}
\usepackage{graphicx}
\usepackage{balance}
\usepackage{multirow}
\usepackage{graphicx}
\usepackage{longtable}
\usepackage{hyperref}
\usepackage{xcolor}

\begin{document}

\title{DTR-Bench: An \textit{in silico} Environment and Benchmark Platform for Reinforcement Learning Based Dynamic Treatment Regime}

\author{Zhiyao Luo,
       Mingcheng Zhu,~\IEEEmembership{Student Member,~IEEE,},
       Fenglin Liu,~\IEEEmembership{Student Member,~IEEE,},
       Jiali Li,~\IEEEmembership{Member,~IEEE,},
       Yangchen Pan,
       Jiandong Zhou,~\IEEEmembership{Member,~IEEE,},
        and~Tingting~Zhu,~\IEEEmembership{Member,~IEEE}
\thanks{Zhiyao Luo, Mingcheng Zhu, Fenglin Liu, Yangchen Pan and Tingting Zhu are with the Department of Engineering Science, University of Oxford, UK}
\thanks{Jiali Li is with the Department of Chemistry, National University of Singapore, Singapore.}
\thanks{Jiandong Zhou is with the Department of Family Medicine and Primary Care, Li Ka Shing Faculty of Medicine, School of Public Health, Department of Pharmacology and Pharmacy, Li Ka Shing Faculty of Medicine, The University of Hong Kong, Hong Kong SAR, China and Division of Health Science, Warwick Medical School, University of Warwick, Coventry, UK.}}

\markboth{UNDER REVIEW}%
{Shell \MakeLowercase{\textit{Luo et al.}}: A Sample Article Using IEEEtran.cls for IEEE Journals}

\maketitle

\begin{abstract}
Reinforcement learning (RL) has garnered increasing recognition for its potential to optimise dynamic treatment regimes (DTRs) in personalised medicine, particularly for drug dosage prescriptions and medication recommendations. However, a significant challenge persists: the absence of a unified framework for simulating diverse healthcare scenarios and a comprehensive analysis to benchmark the effectiveness of RL algorithms within these contexts. To address this gap, we introduce \textit{DTR-Bench}, a benchmarking platform comprising four distinct simulation environments tailored to common DTR applications, including cancer chemotherapy, radiotherapy, glucose management in diabetes, and sepsis treatment. We evaluate various state-of-the-art RL algorithms across these settings, particularly highlighting their performance amidst real-world challenges such as pharmacokinetic/pharmacodynamic (PK/PD) variability, noise, and missing data. Our experiments reveal varying degrees of performance degradation among RL algorithms in the presence of noise and patient variability, with some algorithms failing to converge. Additionally, we observe that using temporal observation representations does not consistently lead to improved performance in DTR settings. Our findings underscore the necessity of developing robust, adaptive RL algorithms capable of effectively managing these complexities to enhance patient-specific healthcare. We have open-sourced our benchmark and code at \url{https://github.com/GilesLuo/DTR-Bench}.
\end{abstract}


\section{Introduction}

\IEEEPARstart{P}{ersonalised} medicine \cite{chan2011personalized} is at the leading edge of modern healthcare, and it involves customising medical treatments and interventions to match the specific traits and needs of an individual patient. This tailored approach is becoming more important as we aim to treat complex diseases effectively that progress differently in each person and elicit a wide spectrum of responses to the same therapy. Personalised medicine aims to move away from a one-size-fits-all model and instead provide care optimised for the individual. Within this context, Dynamic Treatment Regimes (DTRs) \cite{murphy2003optimal, chakraborty2014dynamic} have become a crucial part of Personalised medicine. DTRs provide a framework for tailoring healthcare interventions to each patient's unique and changing needs. By taking into account individuals' characteristics and evolving responses over time, DTRs allow for more nuanced and effective management of diseases. This represents a major improvement over one-size-fits-all approaches to treatment. DTRs move beyond blanket recommendations to enable customised medical care based on the progress and needs of each patient.

Optimisation of DTRs has found a powerful ally in reinforcement learning (RL), a method known for its exceptional ability to adapt and learn in dynamic and complex environments \cite{yu2021reinforcement, coronato2020reinforcement}. This adaptability aligns with the intricacies of personalised healthcare, where patient conditions and treatment responses continually evolve. RL's approach, which emphasises the iterative refinement of strategies through trial and error, ensures that treatment plans remain in harmony with the changing clinical landscapes of patients. In this context, the use of RL suggests a promising direction for the development of efficient DTRs, demonstrating the potential for automated and adaptive healthcare solutions that can operate and adjust in real-time. 

Although integrating RL into DTRs shows promise, it poses significant evaluation challenges. In particular, the ethical and safety concerns of the use of RL require that an algorithm be thoroughly tested \textit{in silico} before any real-world implementation. This highlights the need for rigorous evaluation of RL-based DTRs in simulated environments first to address the ethical constraints and potential risks associated with directly trialling these adaptive algorithms on patients. Thorough \textit{in silico} testing is an essential step to ensure the safety and efficacy of RL-driven personalised medicine approaches.

\begin{figure*}[t]
     \centering
     \includegraphics[width=0.99\textwidth]{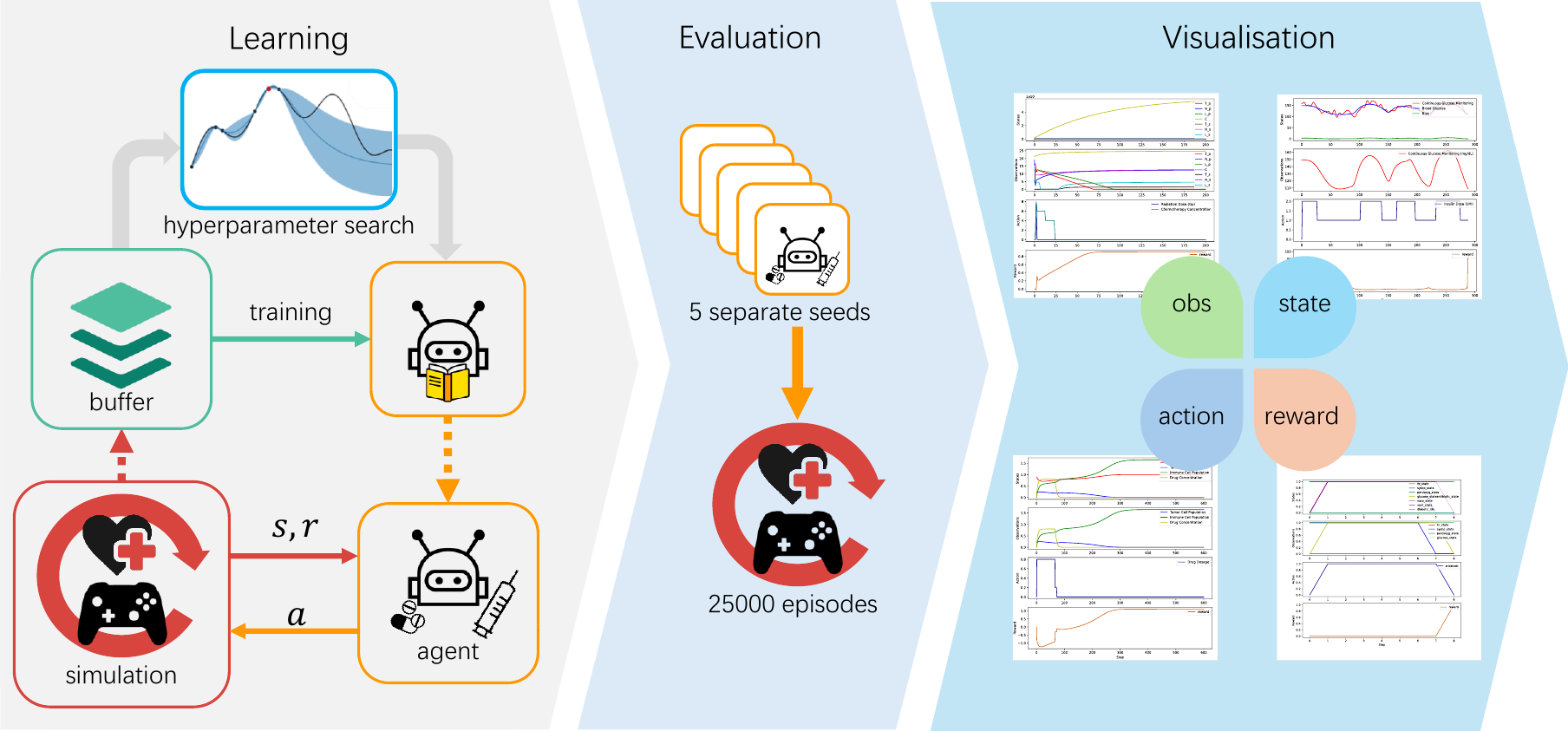}
    \caption{\textbf{Workflow of the DTR-Bench platform.} The platform streamlines 3 steps of \textit{DTR-bench}. \textbf{Step 1 Learning:} RL algorithms interact with the environment, capturing interaction trajectories in a buffer for efficient off-policy learning. Hyperparameters are tuned using Tree-structured Parzen Estimator (TPE) optimisation\cite{ozaki2020multiobjective}. \textbf{Step 2: Evaluation:} Models are retrained with the optimised hyperparameters using five distinct seeds, with each model undergoing testing across 5,000 episodes to ensure a fair assessment. \textbf{Step 3 Visualisation:} The platform facilitates individual and cohort-averaged trajectory visualisations, supporting intuitive model development and analysis.}
    \label{fig:three sin x}
\end{figure*}

A commonly used approach is off-policy evaluation (OPE) \cite{thomas2016data, uehara2022review}, which evaluates the performance of an RL algorithm without new data or experiments. However, OPE's reliability is compromised by issues such as insufficient data coverage and the limited availability of samples, which are problems exacerbated by the imbalanced nature of medical retrospective data. For example, a study\cite{gottesman2018evaluating} raised concerns about the diverse evaluation outcomes across OPE methods on a Sepsis treatment recommendation problem, showcasing that the imbalanced data might increase evaluation variance. It is also reported that the quality of behavioural policy under imbalanced data is important\cite{raghu2018behaviour}, further calling for reconsidering RL evaluation in DTR problems. Therefore, the complexity of real-world testing and the challenges associated with OPE in healthcare data highlight the broader difficulties of implementing RL in DTR. 

Simulation modelling has emerged as an essential tool in response to the difficulties of directly testing RL-based DTRs in clinical settings. By leveraging approaches such as ordinary differential equations (ODEs) and structural causal 
 models (SCMs), simulations can create rich, detailed DTR scenarios based on extensive medical knowledge and patient data. Combined with RL algorithms, these models offer a promising avenue for evaluating the performance of treatment regimes without exposing patients to risk. This methodology has been applied across various medical domains, including cancer chemotherapy \cite{ribba2012tumor} and chronic conditions management like diabetes\cite{nasir2022population}, demonstrating the utility of simulation in developing and refining DTRs. Apart from mathematical modelling, neural networks (NNs), including generative adversarial networks, have also been utilised to simulate clinical data \cite{yoon2019time,brophy2023generative}, offering another viable path for creating realistic patient datasets for research. 

Despite their potential, NN approaches can suffer from data coverage, learning bias and out-of-distribution error under medical data imbalance\cite{douzas2018effective}. This limitation highlights the inherent uncertainty and associated risks when employing NN approaches as environment simulators for RL experiments. While simulation models are an improvement for evaluating DTRs, the simplified simulations raise valid concerns about whether model findings will translate accurately to actual clinical settings. To bridge this gap, simulations should more closely mimic the nuances of patient conditions, treatment responses, and outcomes. Mathematical modelling in disease treatment, typically characterised by ODEs and SCMs, faces inherent limitations: 
\begin{itemize}
    \item \textbf{Noise}. Traditional simulation models often overlook noise in observations, a critical aspect of real-world data; 
    \item \textbf{PK/PD Variance}. Existing models generally use a uniform set of parameters for all patients, neglecting individual differences. This uniformity raises questions about whether RL algorithms genuinely understand treatment principles or merely memorise patterns to achieve high rewards in the simulation environment, as there is no account for individual variability; 
    \item \textbf{Hidden Variables and Missing Values}. Real medical data often includes missing values and inaccessible variables due to clinical decisions or the invasive nature of data collection.
\end{itemize}

Addressing these issues is crucial for creating simulations that better reflect the complexity of real-life healthcare situations. Historically, many studies have overlooked these important details, resulting in a disconnect between the applicability and reliability of simulation results.

\begin{figure*}[ht]
     \centering
     \begin{subfigure}[b]{0.38\textwidth}
         \centering
         \includegraphics[width=\textwidth]{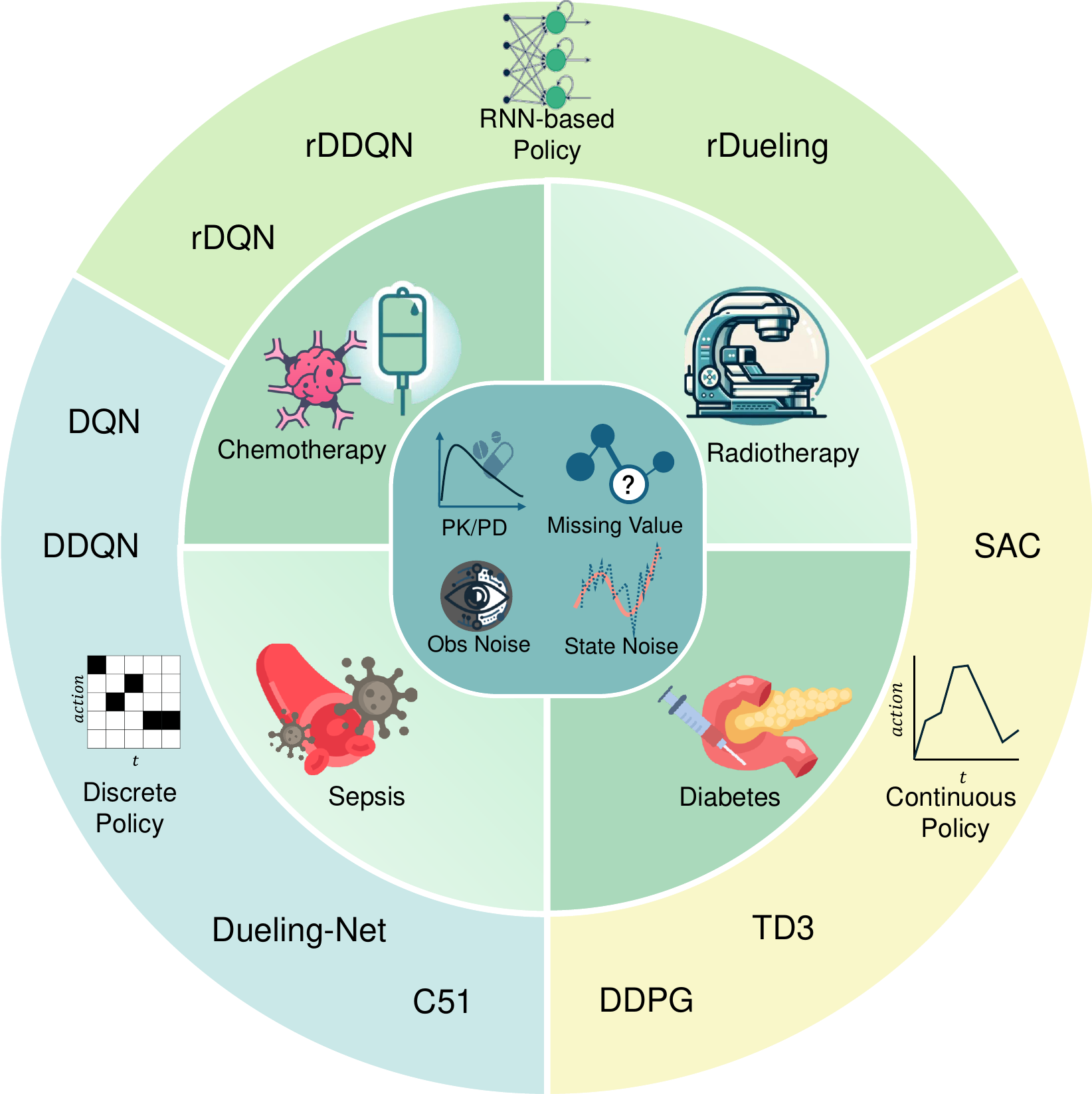}
         \caption{ }
     \end{subfigure}
     \hfill
          \begin{subfigure}[b]{0.46\textwidth}
         \centering
         \includegraphics[width=\textwidth]{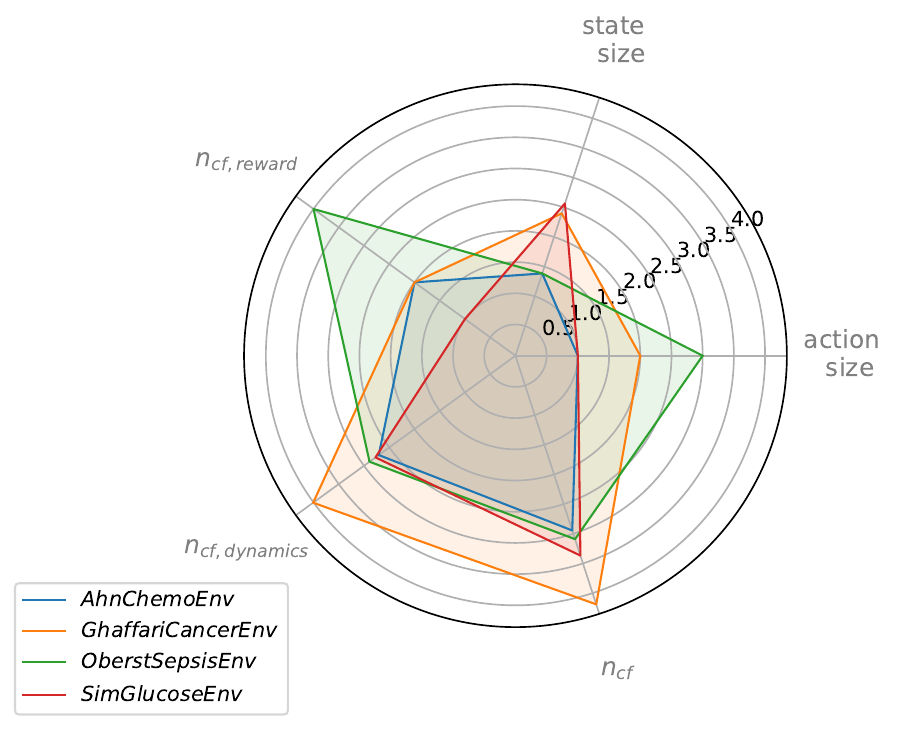}
         \caption{ }
     \end{subfigure}
     \caption{\textbf{A  summary of  RL algorithms and environments in the DTR-Bench platform.} \textbf{a)} DTR-Bench benchmarks discrete-action, continuous-action, and sequential RL algorithms—across four critical healthcare challenges: chemotherapy, radiotherapy, sepsis, and Type-1 diabetes management. The platform evaluates these algorithms rigorously by considering complex factors such as PK/PD modelling, missing values, and noise in observations and states. \textbf{b)}A radar plot showing the environment configurations, where $n_{\text{cf}}$ means the counterfactual variables, $n_{\text{cf, reward}}$ and $n_{\text{cf, dynamics}}$ mean the counterfactual variables affecting the reward function and patient PK/PD dynamics, respectively. $n_{\text{cf}}$, $n_{\text{cf, reward}}$ and $n_{\text{cf, dynamics}}$ are with Logarithm}
     \label{fig: pipeline and radar plot}
\end{figure*}
In this work, we introduce \textit{DTR-Bench}, a growing in silico RL simulation platform developed specifically towards more realistic DTR environment simulators. The design of \textit{DTR-Bench} is grounded in a commitment to closely replicate the nuanced intricacies of healthcare scenarios, thereby providing a robust framework for exploring and evaluating RL algorithms. First, we incorporate noise in observations, pharmacokinetic/pharmacodynamic (PK/PD) variance, hidden variables, and the reality of missing data into our simulation design. This approach aims to enhance the realism of the simulated environments and, by extension, the validity of evaluating RL algorithms in these more nuanced and variable settings. Then, we present a comprehensive benchmarking analysis that leverages the capabilities of \textit{DTR-Bench} across four medical simulation environments, including the scheduling of chemotherapy drugs for cancer treatment, the optimisation of tumour growth and treatment strategies, the management of glucose levels in diabetic patients, and the formulation of sepsis treatment protocols. Our contributions include:
\begin{enumerate}
    \item Developing a comprehensive pipeline that integrates practical clinical considerations into simulation experiment setups.
    
    \item Implementing a collection of mathematical simulation environments for benchmarking RL algorithms, a first in the field of DTR.

    \item Standardising RL hyperparameter tuning and evaluation within a treatment regime context, offering a unified and replicable methodology for algorithm evaluation.
    
    \item Establishing the benchmark in DTRs across various clinical research areas, laying the groundwork for broad, interdisciplinary applications as a first in the DTR field.
\end{enumerate}

\section{Related Work}
\subsection{RL-based Dynamic Treatment Regime and Benchmarks}
RL applications in DTRs have been divided into two main branches: simulation-based and real-world data-based approaches. Each offers unique insights and challenges.

\textbf{Real-World Data-based DTRs} primarily leverage observational healthcare data to train and evaluate RL models. The reliance on off-policy evaluation \cite{dann2014policy, voloshin2019empirical,tang2021model}, such as in \cite{tang2022leveraging, komorowski2018artificial,wu2023value}'s work, is crucial here due to the ethical and logistical challenges associated with deploying experimental policies directly in a clinical setting.  However, the absence of online testing in real-world data-based DTR makes it difficult to validate and iteratively improve RL algorithms under true clinical conditions. This limitation often results in a gap between the theoretical improvement these algorithms can provide and their proven clinical effectiveness.

\textbf{Simulation-based DTRs} allow for the controlled testing of RL algorithms in varied healthcare scenarios without the ethical concerns of direct patient intervention, such as \cite{zhu2020basal, fox2020deep, bhattarai2023using}. Simulation-based DTRs benefit from the ability to conduct exhaustive testing across a wide range of hypothetical scenarios. For instance, an algorithm can perform unlimited trial-and-error iterations on a virtual patient, a practice that is not feasible in real-world scenarios.

\subsection{Benchmarks for RL algorithms}
Despite the extensive benchmarks established for RL in domains such as games\cite{badia2020agent57, gulcehre2020rl} and autonomous driving\cite{jiang2023reinforcement}, a similar benchmark in DTRs has been conspicuously lacking. This gap highlights a significant need in the field to provide a robust platform that can systematically evaluate and compare the performance of RL algorithms in complex, dynamic healthcare settings. \textit{DTR-Bench} aims to fill this gap by offering a standardised, diverse set of simulation environments that mimic real-world treatment dynamics and patient responses, providing a critical tool for advancing RL research in healthcare.

\subsection{On-policy and Off-policy RL}
Reinforcement Learning (RL) constitutes a pivotal branch of machine learning, concentrating on cultivating agents capable of sequential decision-making to optimise cumulative rewards within specified environments. These agents navigate through uncertain and complex settings, deriving knowledge from the feedback of their actions to achieve predefined objectives. This section elucidates foundational and advanced RL algorithms catered to discrete and continuous action spaces, thereby setting the stage for their benchmarking in later discussions.

\textbf{On-policy vs. Off-policy Learning:} RL methodologies bifurcate into on-policy and off-policy paradigms. On-policy methods require agents to adhere strictly to the currently evaluated policy during learning and decision-making processes. Conversely, off-policy strategies allow for evaluating a policy different from the one being followed, offering flexibility in learning from actions not taken by the current policy. This flexibility is paramount in healthcare settings where treatment decisions might deviate from AI recommendations, reflecting a blend of clinical expertise and AI-driven insights. The retrospective nature of medical data collection further underscores the need for off-policy approaches, accommodating the temporal gaps in policy updates. Therefore, we only consider off-policy learning in this paper.

\textbf{Value-based, Policy-based, and Actor-Critic Methods:} At the core of RL lie three principal strategies: value-based, policy-based, and actor-critic methods. Value-based approaches focus on maximising a value function to determine the best action from any state. Policy-based methods directly learn the policy that maps states to actions, optimising for long-term rewards. Policy-based algorithms are omitted in this work as they must be on policy. The actor-critic framework combines value-based and policy-based approaches, featuring an actor who suggests actions based on the current policy and a critic who evaluates these actions by estimating the value function. The actor-critic framework facilitates efficient learning by leveraging the strengths of both value approximation and policy optimisation.

\subsection{Algorithms for Discrete Action Spaces}

\textbf{Deep Q-Network (DQN)\cite{mnih2015human}:} DQN integrates deep neural networks with Q-learning, a value-based RL algorithm. The key innovation of DQN is to use a deep neural network to approximate the Q-value function, which estimates the value of taking an action in a given state. The loss function for DQN is defined as:
\begin{equation}
\resizebox{.9\hsize}{!}{$
    L_{\text{DQN}}(\theta) = \mathbb{E}\left[\left(r + \gamma \max_{a'}Q\left(s', a'; \theta^-\right) - Q(s, a; \theta)\right)^2\right]
$}
\end{equation}
where \( \theta \) represents the parameters of the Q-network, \( \theta^- \) the parameters of a target network used for stability, \( s \) and \( s' \) the current and next state, \( a \) and \( a' \) the current and next action, and \( r \) the reward. DQN introduced experience replay and fixed Q-targets to stabilise training.

\textbf{Double Deep Q-Network (Double DQN)\cite{van2016deep}:} Double DQN addresses the overestimation bias of Q-values in DQN. It decouples the selection and evaluation of the action in the Q-update target ($Y$):
\begin{equation}
\resizebox{.9\hsize}{!}{$
\begin{aligned}
      L_{\text{DDQN}}(\theta) &= \mathbb{E}\Bigg[\Bigg(r + \gamma Q\Big(s', \underset{a'}{\mathrm{argmax}} Q\left(s', a'; \theta\right); \theta^-\Big)\\
       &- Q\left(s, a; \theta\right)\Bigg)^2\Bigg],
\end{aligned}
$}
\end{equation}
This modification reduces overestimation by using two sets of weights, one for selecting the best action and another from the target network to evaluate its value.

\textbf{Dueling Network Architecture\cite{wang2016dueling}:} The Dueling Network decouples value and advantage streams within the network architecture, providing a more robust estimate of the state value while maintaining the advantage of each action:
\begin{equation}
\resizebox{.75\hsize}{!}{
$
    Q(s, a) = V(s) + A(s, a) - \frac{1}{|\mathcal{A}|} \sum_{a'}A(s, a')
$}
\end{equation}
where \( V(s) \) represents the value function, and \( A(s, a) \) is the advantage function, indicating the relative importance of each action.

\textbf{Categorical DQN (C51)\cite{bellemare2017distributional}}: Categorical DQN enhances the traditional DQN framework by representing the Q-value function as a distribution over a set of discrete outcomes rather than a single expectation. This approach enables a more nuanced understanding of the value distribution, capturing the intrinsic uncertainty of returns. The Q-values in C51 are represented by a distribution of \(Z\) values over a set of predefined supports \(z\) (outcome values), which are discretised into \(N\) bins. The key formula that distinguishes C51 from DQN is its projection of the Bellman update onto a discrete distribution, defined as:
\begin{equation}
\begin{aligned}
        \text{Pr}(Z_{s, a} = z_i) &= \sum_{j=1}^{N} \text{Pr}(Z_{s', a^*} = z_j) \cdot \text{Pr}(R = r | s, a) \\
        & \quad \cdot \mathbb{I}[z_i = r + \gamma z_j],
\end{aligned}
\end{equation}

where \(Z_{s, a}\) is the random variable that represents the return distribution for taking action \(a\) in the state \(s\), and \(\mathbb{I}\) is the indicator function, which is 1 if the condition is true and 0 otherwise. This projection operation essentially redistributes the probability mass of the predicted next state and reward distribution back onto the predefined set of supports, taking into account the Bellman equation with a discrete twist.

\subsection{Algorithms for Continuous Action Spaces}

\textbf{Deep Deterministic Policy Gradient (DDPG)\cite{lillicrap2015continuous}:} DDPG is a model-free, off-policy actor-critic algorithm using deep function approximators that can operate over continuous action spaces. It combines the concepts of DQN and policy gradient methods. The critic updates its parameters by minimising the loss:
\begin{equation}
\resizebox{.9\hsize}{!}{
$
    L_{\text{DDPG}}(\theta^Q) = \mathbb{E}\left[\left(r + \gamma Q'\left(s', \mu'(s'; \theta^{\mu'}); \theta^{Q'}\right) - Q(s, a; \theta^Q)\right)^2\right]
$}
\end{equation}
where \( \mu \) denotes the actor network, $\theta^Q$ and  $\theta^{\mu}$ are the parameters for the critic and actor network, respectively.

\textbf{Twin Delayed DDPG (TD3)\cite{fujimoto2018addressing}:} TD3 improves upon DDPG by addressing its function approximation errors. It introduces three key techniques: clipped double-Q learning, delayed policy updates, and target policy smoothing. These techniques collectively reduce overestimation bias and stabilise the training process.

\textbf{Soft Actor-Critic (SAC)\cite{haarnoja2018soft}:} SAC is an off-policy actor-critic algorithm based on the maximum entropy RL framework, designed for continuous action spaces. It aims to maximise both the expected return and entropy of the policy, leading to more robust and diverse behaviours.

SAC employs two Q-functions, \( Q_{\theta_1} \) and \( Q_{\theta_2} \), to mitigate the overestimation bias common in value-based methods. The parameters of these Q-functions are updated by minimising the mean squared error between the current Q-values and the target Bellman values. The loss function for each Q-function is defined as:
\begin{equation}
\resizebox{.9\hsize}{!}{
$
\begin{aligned}
       L_{\text{SAC}}(\theta_i) &= \mathbb{E}_\mathcal{D}\Bigg[ \Big(Q_{\theta_i}(s, a) - \\
       &\Big(r + \gamma \big(\min_{j=1,2} Q_{\theta'_j}(s', a') - \alpha \log \pi_\phi(a'|s')\big)\Big) \Big)^2 \Bigg],
\end{aligned}
$
}
\end{equation}

where \( \theta'_j \) represents the parameters of the target Q-networks, and \(\mathcal{D}\) is the experience replay buffer.

The policy \(\pi_\phi\) is trained to maximise a trade-off between expected Q-values and the entropy term, which facilitates exploration. The policy objective function is:
\begin{equation}
\resizebox{.9\hsize}{!}{$
   J_{\text{SAC}}(\phi) = \mathbb{E}_{s \sim \mathcal{D}, a \sim \pi_\phi}\left[ \alpha \log (\pi_\phi(a|s)) - \min_{i=1,2} Q_{\theta_i}(s, a) \right].$}
\end{equation}

\section{Methods}
\label{section:methods}
\subsection{Problem Formulation}
 A DTR problem is formulated as a Partially Observable Markov Decision Process (POMDP) with a set of parameters \( \mathcal{P}=\{\mathcal{S}, \mathcal{A}, p, R, \mathcal{O}, \gamma\} \). Here, \( \mathcal{S} \) is the state space; \( \mathcal{A} \) is the action space for the treatment decision; \( p(s_{t+1}|s_t, a_t) \) is the transition function that maps the transition from the current state \( s_t \) to the next state \( s_{t+1} \) based on action \( a_t \). \( R: \mathcal{S} \times \mathcal{A} \rightarrow \mathbb{R} \) is the reward function that assigns values to state-action pairs, and \( \gamma \in (0, 1) \) is the discount factor, emphasising the importance of immediate rewards over future ones. In the POMDP setting, the agent does not have direct access to the true state but receives an observation \( o \in \mathcal{O} \), where \( \mathcal{O}: \mathcal{S} \times \mathcal{A} \rightarrow \mathcal{O} \) is the observation model. This observation is generated from the underlying state according to a probability distribution \( o \sim \mathcal{O}(s) \). The objective of the RL agent is to discover an optimal policy \( \pi: \mathcal{O} \rightarrow \mathcal{A} \) that maximises the expected cumulative reward, defined as \( G(s_t) = \mathbb{E}_{\pi}[\sum_{t}{\gamma^{t-1}R(s_t, a_t)}] \).

 In DTR, the transition function $p$ represents the treatment response dynamics of patients, which is unavailable. To test the treatment regime algorithms in silicon before the implementation in the real world, a simulated transition $\hat{p}$  is used to approximate the real patient response. Ideally, we expect 
\[
\forall s \in \mathcal{S}, \forall a \in \mathcal{A}, \quad \hat{p}(s'|s, a) \approx p(s'|s, a)
\]

\subsection{Enhancement towards more realistic DTR Simulation}
To bridge the gap between conventional DTR simulations and the complexities of real-world scenarios, we identify the discrepancies that often make simulations overly simplistic and detached from clinical reality. Our enhancements are motivated by the recognition that simulations often fail to capture the multifaceted nature of real-world scenarios. Traditional DTR simulations, if without noise or dynamic variation, risk producing models that excel in memorisation rather than generalisation. Such models are prone to overfitting and cannot perform well across the diverse situations encountered in clinical practice. Therefore, It is essential to introduce variations and noises into the simulation environment to train a DTR agent that is more realistic and generalisable. Hence, we identify the following enhancements in each simulation environment:

\begin{itemize}
    \item \textbf{Partial Observability.} In real-world healthcare scenarios, not all patient information is readily measurable or available. To simulate this, our model incorporates partial observability, where certain state variables are intentionally hidden or obscured. This reflects the practical limitations of medical diagnostics and acknowledges that treatment decisions often must be made with incomplete information. It challenges the RL algorithms to make effective decisions based on limited data, mirroring real-life clinical situations.
    \item \textbf{PK/PD Variation.} Pharmacokinetics and Pharmacodynamics variations are critical in modelling how different groups of patients respond to medications. To replicate this, patient parameters in our simulations are initialised randomly within a defined range. This variability represents the diverse responses of patients to drugs, ensuring that our model captures the heterogeneity seen in actual patient populations. This aspect tests the adaptability of RL algorithms in tailoring treatment strategies to varied patient profiles.
    \item \textbf{Observation Noise.} Observational noise is an inherent part of medical data, stemming from measurement inaccuracies and systemic errors. Our model simulates this by adding noise to the observed states. This addition of noise ensures that the RL algorithms are trained and tested in conditions that closely resemble the uncertainty present in real-world medical data. 
    \item \textbf{Missing Values. }Missing values are a common challenge in medical datasets, arising due to various reasons such as the invasiveness of certain tests or clinical judgments deeming certain data collection unnecessary. To mimic this, our simulation randomly masks out a portion of the observations, reflecting a fixed ratio of missing data. This setup evaluates the resilience and robustness of RL algorithms in dealing with incomplete information, a common occurrence in clinical settings.

\end{itemize}


\subsection{Learning and Evaluation Workflow}

To address the critical role of hyperparameters in the performance of RL algorithms, we developed a model-agnostic standardised train-tuning-evaluation pipeline. This evaluation scheme ensures consistent and equitable assessments of RL policies by encompassing the entire process, from agent training and hyperparameter tuning to policy evaluation, as depicted in the accompanying illustration (see Figure \ref{fig: pipeline and radar plot}). 

In RL, where hyperparameters significantly influence outcomes, our framework mandates a set of uniform hyperparameters across all algorithms to facilitate fair comparisons. We adopt a fixed random seed for the initial stages of training and hyperparameter optimisation to maintain consistency while employing five diverse seeds during the evaluation phase to rigorously test the model's ability to generalise and prevent overfitting specific seed-induced data scenarios. The choice of hyperparameters is given in Table \ref{tab:hyperparameters}.

Hyperparameters are refined using the Tree-structured Parzen Estimator (TPE) algorithm in the grid space \cite{ozaki2020multiobjective}. TPE estimates the effectiveness of configurations based on previous outcomes, intelligently focusing on promising areas to enhance the probability of identifying optimal settings. For each algorithm, the TPE optimiser is initialised with 50 trials of random search and then optimised up to another 50 trials until hyperparameter search pruning or reach-of-length termination.

\section{Experiment Settings}
In this section, we introduce the key environmental settings critical to developing more realistic Dynamic Treatment Regime (DTR) simulations. We aim to bring overly simplified models closer to real-world medical decision-making processes. We experimented with four critical components in designing DTR simulators (i.e., hidden variables, PK/PD, noise, and missing values) and their consequence towards RL performance on various distinct treatment regime scenarios. We detail the fundamental structure of each DTR environment and its problem setups in reinforcement learning (RL) terms, including observation space, action space, and reward design.

Our study analyses four simulated medical environments, each targeting a specific disease and type of treatment. These environments are \textit{AhnChemo} \cite{ahn2011drug}, \textit{OberstSepsis} \cite{oberst2019counterfactual}, \textit{GhaffariCancer} \cite{ghaffari2016mixed}, and \textit{SimGlucose} \cite{man2014uva} \footnote{In the case that the environment is without its original name, we name the environment as the initial of the first author and context.} They have been chosen for their relevance to different diseases and various treatment approaches, providing a comprehensive range of scenarios to test the RL algorithms. Table \ref{tab:env summary} summarises each environment's description.

\begin{table*}[htbp]
    \small
    \caption{\textbf{Summary of Simulated Medical Environments}. This table provides an overview of the four simulated environments used in our study, highlighting the targeted disease, the type of treatment, the dynamic system model, and the nature of the action space for each environment. Cont. means continuous, Disc. means discrete, and Cont./Disc. means that the environment is natively continuous but adaptable to discrete action space by action binning. }
    \label{tab:env summary}
    \centering
    \begin{tabular}{c|c|c|c|c}
    \hline
    Environment & Disease &  Treatment & Dynamics& Action Space \\
    \hline
      \textit{AhnChemoEnv}\cite{ahn2011drug} & Cancer & Chemotherapy & ODE& Cont./Disc. \\
      \textit{GhaffariCancerEnv}\cite{ghaffari2016mixed} & Cancer & Chemotherapy \& Radiotherapy & ODE& Cont./Disc. \\
      \textit{OberstSepsisEnv}\cite{oberst2019counterfactual} & Sepsis & Antibiotics, Mechanical Ventilation, Vasopressors & SCM& Disc. \\
    \textit{SimGlucoseEnv}\cite{man2014uva} & Type-1 Diabetes & Insulin Administration & ODE& Cont./Disc. \\
    \hline       
    \end{tabular}
\end{table*}
Each environment is carefully designed to simulate the complexities and dynamics of real-world medical scenarios. They incorporate varying levels of detail and sophistication in their models, ranging from ordinary differential equations (ODEs) to structural causal models. This variety ensures a robust and comprehensive evaluation of the RL algorithms across different healthcare contexts.

The details of the design of each simulated environment, including the disease dynamics, observation space, action space, and reward functions, are described in the following content.

\subsection{\textbf{\textit{AhnChemoEnv}: A Comprehensive Chemotherapy Simulation Model}}

The \textit{AhnChemoEnv} is an advanced simulation environment designed for modelling cellular dynamics under the influence of chemotherapy \cite{ahn2011drug}. Chemotherapy is a way to treat cancer by using medicine to kill cancer cells. This treatment can stop or slow down the growth of tumours, but it can also harm healthy cells because it doesn't only target cancer cells. The \textit{AhnChemoEnv} incorporates interactions between tumour cells, normal cells, immune cells, and factors in the impact of chemotherapy. 

\textbf{Dynamics Formulation:}
The dynamics in chemotherapy are modelled by a system of ordinary differential equations (ODEs). The ODEs are built upon established mathematical frameworks and biological assumptions reported in \cite{kuznetsov1994nonlinear}. The immune response, competition terms, and control theory for chemotherapy are grounded in known biological interactions, including the immune system's ability to recognize and fight tumour cells, the competition for resources between tumour and normal cells, and the dynamics of drug treatment efficacy and toxicity \cite{de2003dynamics}. The ODEs can be expressed by

\begin{equation}
\resizebox{.9\hsize}{!}{
$
    \left\{
    \begin{aligned}
       &\frac{dN}{dt} = r_2N(1-b_2N) - c_4TN - a_3(q-e^{-B})N,\\
       &\frac{dT}{dt} = r_1T(1-b_1T) - c_2IT - c_3TN - a_2(1-e^{-B})T,\\ 
       &\frac{dI}{dt} = s + \frac{\rho IT}{\alpha+T} - c_1 IT - c_3 TN - a_1(1-e^{-B})T,\\
       &\frac{dB}{dt} = -d_2B + u(t)
    \end{aligned}
    \right.
$
}
\end{equation}

The first equation simulates normal cell growth and loss due to tumour competition; the second equation models tumour cell behaviour, including growth and death from interactions with normal and immune cells; the third equation describes changes in the immune cell population, influenced by external sources, tumour presence, competition, and natural death; and the final part focuses on the decay of drug concentration in the blood and the injection of new drugs. The time unit for this ODE system is day, and the default time interval between steps is 6 hours.

\textbf{Variables Description:} The variables within the \textit{AhnChemoEnv}  play specific roles in the simulation. A summary of these variables is provided in Table \ref{table:ahn_variables}.

\begin{table*}[ht]
\small
    \caption{Variables of the \textit{AhnChemoEnv} ODEs system}
    \label{table:ahn_variables}
    \centering
    \begin{tabular}{c c p{5cm} c c}
    \hline
    \textbf{Variable Name} & \textbf{Usage} & \textbf{Description} & \textbf{Unit} & \textbf{Range} \\ 
    \hline
       $N(t)$ & $\mathcal{S}$& Normal cell population & No. cells ($\times 10^{11}$)& (0, 2)\\ 
       $T(t)$ & $\mathcal{O}$& Tumour cell population, representing tumour burden & No. cells ($\times 10^{11}$)& $(0, 2)$ \\ 
       $I(t)$ & $\mathcal{O}$& Immune cell population, measuring immune response level & No. cells ($\times 10^{11}$)& $(0, 2)$ \\ 
       $B(t)$ & $\mathcal{O}$& Drug concentration in the bloodstream & unit/L & $(0, 1)$ \\ 
       $u(t)$ & $\mathcal{A}$& Chemotherapy drug administration rate & unit & $(0, 1)$ \\ 
    \hline
    \end{tabular}
\end{table*}

The observation space $\mathcal{O}$ comprises the tumour population $T(t)$, the immune population $I(t)$, and the drug concentration $B(t)$. The normal cell population $N(t)$ is hidden as normal cells are not commonly accessible during treatment or diagnosis. The cell counts have been scaled down by a factor of $10^{11}$ so that one unit represents the carrying capacity of the normal cells in the region of the tumour. It can also prevent overflow problems due to excessively large values. The agent should control $u(t)$, the levels of chemotherapy dosage to eliminate tumour cells in a minimum time with minimal harm to normal cells. 

\textbf{Parameters Description:} The parameter values of the ODEs system are determined based on clinical knowledge \cite{de2003dynamics}. The descriptions and values are shown in Table \ref{table:ahn_parameters}.

\textbf{Reward Function:} The reward $R$ is defined as

\begin{equation}
    \label{ahn:reward}
    r(t) = \frac{N(t)}{N(0)}-\frac{T(t)}{T(0)} + I(t) - u(t)
\end{equation}

It is designed to balance several key factors in cancer treatment: it encourages the increase of normal cells and immune cells, as indicated by the terms \( \frac{N(t)}{N(0)} \) and \( I(t) \), respectively. Currently, it aims to reduce the number of tumour cells, as shown by the negative term \( -\frac{T(t)}{T(0)} \). Additionally, the function encourages favourable outcomes with minimal drug dosage, denoted by \( -u(t) \). This aspect underscores the importance of reducing drug usage to possibly reduce side effects and costs.

\subsection{\textbf{\textit{GhaffariCancerEnv}: A Mixed Radiotherapy and Chemotherapy Model}}

The \textit{GhaffariCancerEnv} represents a comprehensive simulation environment that models the interactions between normal cells, cancer cells, radiotherapy, and chemotherapy agents at different sites, incorporating the metastatic spread of cancer from a primary to a secondary site \cite{ghaffari2016mixed}. It details how radiotherapy, a treatment that uses high doses of radiation to eliminate cancer cells and reduce tumour size, and chemotherapy, which uses drugs to destroy cancer cells by inhibiting their growth and division, affect the cellular landscape. Additionally, the model accounts for the time-delayed migration of cancer cells from the primary to the secondary site, adding a crucial dimension to understanding cancer's progression and treatment impacts.

\textbf{Dynamics Formulation:} The dynamics of tumour cell growth and the effect of chemotherapy are determined in studies conducted with mice, alongside established clinical insights into immune cell reactions to cancer \cite{de2003mathematical, de2006mixed}. The influence of radiotherapy and the behaviour of cancer metastasis is determined through mice data in \cite{ghaffari2016mixed}. The dynamics are formulated in ODEs, which can be expressed by

\begin{equation}
\resizebox{.9\hsize}{!}{
$
\left\{
    \begin{aligned}
    \frac{dT_p}{dt} &= a_1 T_p (1 - b_1 T_p) - c_1 N_p T_p - D_p T_p - D T_p\\
    & + \gamma u - K_{T1} \frac{T_p M}{W_{T1} + T_p} \\
    \frac{dN_p}{dt} &= e_1 C - p_1 N_p T_p - f_1 N_p - \epsilon D N_p + \gamma_2 v \\
    & - K_{1N} \left( \frac{N_p M}{W_{1N} + N_p} \right) \\
    \frac{dL_p}{dt} &= -mL_p + j_1 \frac{T_p}{k_1 + T_p} - q_1L_p T_p + r_{11}N_p T_p + r_{12}C T_p \\
    & - u_1 N_p L_p^2 - \epsilon D L_p + \gamma_3 x - K_{1L} \left( \frac{L_p M}{W_{1L} + L_p} \right) \\
    \frac{dC}{dt} &= \alpha - \beta C - K_{1C} \left( \frac{CM}{W_{1C} + C} \right) \\
    \frac{dT_s}{dt} &= a_2 T_s (1 - b_2 T_s) - c_2 N_s T_s - D_s T_s + \alpha_2 T_p (t - \tau) \\
    &- K_{T} \left( \frac{T_s M}{W_{2T} + T_s} \right) \\
    \frac{dN_s}{dt} &= e_2 C - p_2 N_s T_s - f_2 N_s - K_{2N} \left( \frac{N_s M}{W_{2N} + N_s} \right)\\
    \frac{dL_s}{dt} &= -m_2 L_s + j_2 \frac{T_s}{k_2 + T_s} - q_2 L_s T_s + r_{21} N_s T_s \\
    & + r_{22} C T_s - u_2 N_s L_s^2 - K_{2L} \left( \frac{L_s M}{W_{2L} + L_s} \right) \\
    \frac{dc_1}{dt} &= \mu_{c1} v_M \left(1 - \frac{c_1}{k_{c1}}\right) \\
    \frac{dc_2}{dt} &= \mu_{c2} v_M \left(1 - \frac{c_2}{k_{c2}}\right) \\
    \frac{dM}{dt} &= -\mu_M + v_M \\
    \frac{du}{dt} &= D T_p - \gamma_1 u - \delta u \\
    \frac{dv}{dt} &= \epsilon D N_p - \gamma_2 v - \delta v \\
    \frac{dx}{dt} &= \epsilon D L_p - \gamma_3 x - \delta x
\end{aligned}
\right.
$}
\label{eq:ghaode}
\end{equation}

where $D_p = d_1\frac{L_p^l}{sT_p^l+L_p^l}$ and $D_s = d_2\frac{L_s^l}{sT_s^l+L_s^l}$.

The ODE system presented in Equation \ref{eq:ghaode} describes the complex interactions within the cancer environment under treatment. Specifically, $\frac{dTp}{dt}$ represents the dynamics of the tumour cell population at the primary site, accounting for natural growth, inhibition by natural killer (NK) cells, reduction due to radiotherapy and chemotherapy, and the effect of metastasis. $\frac{dN_p}{dt}$ and $\frac{dL_p}{dt}$ model the dynamics of NK cells and CD8+ T cells at the primary site, respectively, including the effects of chemotherapy, radiotherapy, and their interactions with tumour cells. $\frac{dC}{dt}$ describes the dynamics of lymphocytes, excluding NK cells and CD8+ T cells. The equations for $\frac{dT_s}{dt}$, $\frac{dN_s}{dt}$, and $\frac{dL_s}{dt}$ mirror those at the primary site but apply to the secondary site, emphasizing the role of metastasis. The dynamics of chemotherapy agent concentration in the blood is modelled by $\frac{dM}{dt}$, while $\frac{du}{dt}$, $\frac{dv}{dt}$, and $\frac{dx}{dt}$ capture the populations of cancer cells, NK cells, and CD8+ T cells exposed to radiation, respectively. Each equation integrates the influence of treatment modalities, cellular interactions, and the migration of cells between sites to provide a comprehensive view of the cancer treatment landscape.

\textbf{Variables Description:} The variables of the ODEs system are summarised in Table \ref{table:gha variables}. 

\begin{table*}[htbp]
\small
\centering
\caption{Variables of the \textit{GhaffariCancerEnv} ODEs system}
\label{table:gha variables}
\begin{tabular}{c c p{8cm} c c}
\hline
\textbf{Variable Name} &  \textbf{Usage} &  \textbf{Description} & \textbf{Unit} & \textbf{Range} \\
\hline
\( T_p(t) \) &   $\mathcal{O}$ &The total tumour cell population at the primary site 
& cells & $(0, 10^{11})$\\
\( N_p(t) \) & $\mathcal{O}$ &The concentration of NK cells per litre of blood (cells/L) at the primary site & cells / L & $(0, 10^{10})$\\
\( L_p(t) \) & $\mathcal{O}$ &The concentration of CD8+ T cells per litre of blood (cells/L) at the primary site & cells / L & $(0, 10^{10})$\\
\( C(t) \) & $\mathcal{O}$ &The concentration of lymphocytes per liter of blood (cells/L), not including NK cells and active CD8+T & cells / L & $(0, 10^{11})$\\
\( T_s(t) \) & $\mathcal{O}$ &The total tumour cell population at the secondary site & cells & $(0, 10^{11})$\\
\( N_s(t) \) & $\mathcal{O}$ &The concentration of NK cells per liter of blood (cells/L) at the secondary site & cells / L & $(0, 10^{10})$\\
\( L_s(t) \) & $\mathcal{O}$ &The concentration of CD8+ T cells per litre of blood (cells/L) at the secondary site & cells / L & $(0, 10^{10})$\\
\( M(t) \) & $\mathcal{S}$ &The concentration of chemotherapy agent per litre of blood (mg/L) & mg/L & $(0, 10^{10})$\\
\( u(t) \) & $\mathcal{S}$ &The population of cancer cells that have been exposed to radiation & cells & $(0, 10^{11})$\\
\( v(t) \) & $\mathcal{S}$ &The population of NK cells that have been exposed to radiation & cells & $(0, 10^{11})$\\
\( x(t) \) & $\mathcal{S}$ &The population of CD8+ T cells that have been exposed to radiation & cells & $(0, 10^{11})$\\
$ D$ & $\mathcal{A}$ & The radiation dose administered & Gy & (0, 10) \\
$ v_M(t) $ & $\mathcal{A}$ & The per-litre blood dosage of chemotherapy agents & mg/L & (0,8) \\
\hline
\end{tabular}
\end{table*}

\textbf{Parameters Description:} The parameter values of the ODEs system are determined using both mouse and human parameters \cite{de2006mixed}. The parameters' descriptions and values are shown in Table \ref{table:gha params} and \ref{table:gha params2}.

\textbf{Reward Function:} The reward function designed for this environment comprises a per-step tumour reduction reward $r_T(t)$ and an outcome reward $r_o(t)$.

\textbf{$r_T(t)$}: the reward function encourages the agent to decrease the number of tumour cells in both the primary and secondary sites. A penalty is imposed if the total tumour population exceeds the initial population. In contrast, a reward is given when the agent successfully decreases the tumour population below its initial level. The reward is given by
\begin{equation}
    r_T(t) = 1 - \frac{T_p(t)+T_s(t)}{T_p(0)+T_s(0)}
\end{equation}
 \textbf{$r_o(t)$}: The outcome reward is designed to encourage the complete elimination of tumours and discourage early termination caused by the oversized tumour population. In particular, +100 is given if the tumour population is 0 at both the primary and secondary site; -100 is given if any of the tumour population is higher than $10^{11}$ cells/L; and 0 is given otherwise.
    \begin{equation}
    r(t) = \begin{cases} 
        - 100 & \text{if $T_p(t) > 10^{11}$  or $T_s(t) > 10^{11}$,} \\
        + 100 & \text{if $T_p(t)$ < 1 and $T_s(t)$ < 1,} \\
        0 & \text{otherwise}
        \end{cases}
\end{equation}

The total reward function for this environment is a sum of the previous reward functions, denoted by $ r(t) = r_T(t)+r_o(t)$.

\subsection{\textbf{\textit{OberstSepsisEnv}: A Sepsis Simulator}}

\textit{OberstSepsisEnv} is a synthetic environment designed for sepsis management, based on the framework of a Markov Decision Process (MDP) \cite{oberst2019counterfactual}. This environment incorporates four vital signs: heart rate (hr), blood pressure (bp), oxygen concentration (02), and glucose levels (glu), each categorized into discrete states such as low, normal, and high. It offers three treatment options: antibiotics (abx), vasopressors (vaso), and mechanical ventilation (vent), available for application at each decision point. A critical condition of death is modelled to occur when at least three vital signs are concurrently outside the normal range. Additionally, the model includes a binary variable indicating the presence of diabetes.

\textbf{Dynamics Formulation:} The dynamics of \textit{OberstSepsisEnv} is determined based on the experience of author \cite{oberst2019counterfactual}. The transition probability of each variable is shown in Table.\ref{tab:sepsis dynamics}.

\begin{table*}[ht]
\small
\centering
\caption{Transition probability of the \textit{OberstSepsisEnv}}
\label{tab:sepsis dynamics}
\begin{tabular}{cc|ccc|c c}
\hline
\textbf{Step} & \textbf{Variable} & \textbf{Current} & \textbf{New} & \textbf{Change} & \multicolumn{2}{c}{\textbf{Effect}} \\
\hline
\multirow{4}{*}{1} & \multirow{4}{*}{abx} & \multirow{2}{*}{-} & \multirow{2}{*}{on} & \multirow{2}{*}{-} & hr & H $\rightarrow$ N w.p. 0.5 \\

  &     &  &  &  & bp & H $\rightarrow$ N w.p. 0.5 \\
  \cline{3-7}
& & \multirow{2}{*}{on} & \multirow{2}{*}{off} & \multirow{2}{*}{withdrawn} & hr & N $\rightarrow$ H w.p. 0.1 \\

  &     &  &  &  & bp & N $\rightarrow$ H w.p. 0.5 \\
\hline
\multirow{2}{*}{2} & \multirow{2}{*}{vent} & - & on & - & o2 & L $\rightarrow$ N w.p. 0.7 \\
\cline{3-7}
  &      & on & off & withdrawn & o2 & N $\rightarrow$ L w.p. 0.1 \\
\hline
\multirow{12}{*}{3} & \multirow{12}{*}{vaso} & \multirow{7}{*}{-} & \multirow{7}{*}{on} & \multirow{7}{*}{-} & \multirow{5}{*}{bp} & L $\rightarrow$ N w.p. 0.7 (non-diabetic) \\
  &      &   &   &   &  & N $\rightarrow$ H w.p. 0.7 (non-diabetic) \\
  &      &  &  &  & & L $\rightarrow$ N w.p. 0.5 (diabetic) \\
  &      &   &   &   &  & L $\rightarrow$ H w.p. 0.4 (diabetic) \\
  &      &   &   &   &  & N $\rightarrow$ H w.p. 0.9 (diabetic) \\
  \cline{6-7}
  &      &   &   &   & \multirow{2}{*}{glu} & LL $\rightarrow$ L, L $\rightarrow$ N, N $\rightarrow$ H, \\
  &      &   &   &   & &H $\rightarrow$ HH w.p. 0.5 (diabetic) \\
  \cline{3-7}
  &      & \multirow{4}{*}{on} & \multirow{4}{*}{off} & \multirow{4}{*}{withdrawn} &  \multirow{4}{*}{bp} & N $\rightarrow$ L w.p. 0.1 (non-diabetic) \\
  &      &   &   &   &  & H $\rightarrow$ N w.p. 0.1 (non-diabetic) \\
  &      &   &   &   &  & N $\rightarrow$ L w.p. 0.05 (diabetic) \\
  &      &   &   &   &  & H $\rightarrow$ N w.p. 0.05 (diabetic) \\
\hline
4 & hr &  & \multirow{4}{*}{fluctuate}  &   &\multicolumn{2}{c}{vitals spontaneously fluctuate when not affected} \\
5 & bp &  & &  &  \multicolumn{2}{c}{by treatment (either enabled or withdrawn)} \\
6 & o2 &  & &  & \multicolumn{2}{c}{\textbullet\,the level fluctuates $\pm$1 w.p. 0.1, except:} \\
7 & glu &  &  &  & \multicolumn{2}{c}{\textbullet\,glucose fluctuates $\pm$1 w.p. 0.3 (diabetic)} \\
\hline
\end{tabular}
\end{table*}

The model intricately details how these treatments, when turned on or off, influence the transition of vital signs from states such as low (L), normal (N), high (H), super low (LL), and super high (HH), with respective probabilities. For instance, antibiotic use impacts heart rate and blood pressure, while vasopressors have a more complex effect on blood pressure and glucose levels, particularly distinguishing between diabetic and non-diabetic scenarios. This model also accounts for natural fluctuations in vital signs, both under treatment influence and spontaneously. A patient is discharged only when all vital signs become normal and all treatments have been stopped. Death occurs if 3 or more vitals are abnormal. 

\textbf{Variables Description:} The variable descriptions for the \textit{OberstSepsisEnv} are shown in Table \ref{table:oberstsepsis variables}

\begin{table*}[h]
\small
    \centering
    \caption{Variables of the \textit{OberstSepsisEnv} ODEs}
    \label{table:oberstsepsis variables}
    \begin{tabular}{c c l p{6cm}}
    \hline
        \textbf{Variable Name} & \textbf{Usage} & \textbf{Description} & \textbf{Range}\\
    \hline
        hr & $\mathcal{O}$ & Heart rate & [Low, Normal, High] \\
        bp & $\mathcal{O}$ & Blood pressure& [Low, Normal, High] \\
        o2 & $\mathcal{O}$ & Oxygen concentration & [Low, Normal, High] \\
        glu & $\mathcal{O}$ & Glucose levels &
        [Super Low, Low, Normal, High, Super High] \\
        diabetic & $\mathcal{S}$ & The presence of diabetes &
        [diabetic, non-diabetic] \\
        abx & $\mathcal{A}$ & Antibiotics & [On, Off] \\
        vaso & $\mathcal{A}$ & Vasopressors & [On, Off] \\
        vent & $\mathcal{A}$ & Mechanical ventilation & [On, Off] \\
    \hline
    \end{tabular}
\end{table*}

\textbf{Reward Function:} The reward function designed the $OberstSepsisEnv$ is shown in Equation \ref{eq:sepsis reward}. 

\begin{equation}
\label{eq:sepsis reward}
r(t) =
    \begin{cases} 
    1 & \text{if discharged,} \\
    -1 & \text{if deceased,} \\
    0 & \text{otherwise.}
    \end{cases}
\end{equation}

A reward of +1 is granted for patient survival, transitioning the system into an absorbing state that continuously accumulates rewards. In contrast, a penalty of -1 is imposed for patient mortality. For all other outcomes, no reward or penalty is administered to the agent. This reward structure is designed to encourage the agent to give treatment toward a discharged state while avoiding the patient's death.

\subsection{\textbf{\textit{SimGlucoseEnv}: A glucose-insulin simulating environment}}

The \textit{SimGlucoseEnv} describes the glucose-insulin system during a meal \cite{man2014uva}. The model connects plasma glucose and insulin concentrations with their respective fluxes. It simulates how glucose and insulin levels in the bloodstream are influenced by various processes such as glucose absorption, renal excretion, insulin fluxes, and insulin degradation.

\textbf{Dynamics Formulation}: The dynamics are determined based on real-world data from 300 patients, covering a range of metabolic parameters and demographic characteristics. The dynamics are formulated in ODEs, which are developed based on computational models that simulate interactions between insulin dosing, carbohydrate intake, and glucose metabolism as detailed in \cite{kovatchev2009silico}. The ODEs can be expressed by

\begin{equation}
\resizebox{.9\hsize}{!}{
$
\left\{
\begin{aligned}
   &\frac{dG_p(t)}{dt} = EGP(t) + Ra(t) - U_{ii} - E(t) -k_1 G_p(t) + k_2 G_t(t)  \\
   &\frac{dG_t(t)}{dt} = -U_{id}(t) + k_1 G_p(t) - k_2 G_t(t)  \\
   &\frac{dX(t)}{dt} = -p_{2u}\cdot X(t) + p_{2u} \cdot [I(t)-I_b] \\
    &\frac{dI(t)}{dt} = -k_i\cdot[I'(t)-I(t)] \\
    &\frac{dX^L(t)}{dt} = -k_i[X^L(t) - I'(t)]  \\
    & \frac{dS_{sto}(t)}{dt} = CHO(t) - k_{sto} \cdot S_{sto}(t) \\
    & \frac{dQ_{sto}(t)}{dt} = k_{sto} \cdot S_{sto}(t) - k_{gut} \cdot Q_{sto}(t) \\
    & \frac{dQ_{gut}(t)}{dt}= k_{gut} \cdot Q_{sto}(t) - k_{abs} \cdot Q_{gut}(t) \\
\end{aligned}   
\right.
$
}
\end{equation}

where $Ra(t) = \frac{f \cdot k_{abs} \cdot Q_{gut(t)}}{BW}$, $E(t) = k_{e1}[G_p(t)-k_{e2}]$, $U_{id}(t) = \frac{[V_{m0}+V_{mx}X(t)]G_t(t)}{K_{m0}+G_t(t)}$, and $EGP(t) = k_{p1} - k_{p2}G_p(t) - k_{p3}X^L(t)$

This ODE system models glucose absorption (\(Ra(t)\)) from ingested carbohydrates (\(CHO(t)\)), the body's glucose production (\(EGP(t)\)), the dynamics of insulin (\(I(t)\)), and insulin's impact on glucose utilisation (\(X(t)\)) and its delayed action in the liver (\(X^L(t)\)). The equations track glucose concentrations in plasma (\(G_p(t)\)) and tissue (\(G_t(t)\)), account for renal glucose excretion (\(E(t)\)), and quantify insulin-dependent glucose utilisation (\(U_{id}(t)\)). In addition, the model delineates the digestion process, distinguishing between the solid (\(S_{sto}(t)\)) and liquid (\(Q_{sto}(t)\)) carbohydrate states in the stomach before their absorption in the gut (\(Q_{gut}(t)\)). The model directly correlates dietary intake and insulin administration with blood glucose levels through these dynamics, offering a sophisticated tool to simulate glucose-insulin interactions and aiding effective diabetes management strategies.

The criteria for termi-ting or truncating this environment are as follows: \textit{termination} occurs if the basal plasma glucose level $G_p$  falls below 10 or exceeds 600. If neither condition is met, the environment proceeds for 288 steps, equating to a 24-hour treatment period. Upon completion of these 288 steps, the environment is considered \textit{truncated}.

\textbf{Variable Description}: The variable descriptions for the \textit{SimGlucoseEnv} are shown in Table.\ref{table:simglucose variables}.

\begin{table*}[h]
    \small
    \centering
    \caption{Variables of the \textit{SimGlucoseEnv} ODEs}
    \label{table:simglucose variables}
    \begin{tabular}{c c p{8cm} c c}
    \hline
        \textbf{Variable -me} & \textbf{Usage} & \textbf{Description} & \textbf{Unit} & \textbf{Range}\\
    \hline
        $G_p(t)$ & $\mathcal{O}$ & The amount of glucose in plasma & mg/dL & (10, 600) \\
        $G_t(t)$ & $\mathcal{S}$ &The amount of glucose in the tissue & mg/dL & -\\ 
        $I(t)$ & $\mathcal{S}$ & The insulin concentration & U/day & -\\
        $X(t)$ & $\mathcal{S}$ & The insulin action on glucose utilization & - & -\\
        $X^L(t)$ & $\mathcal{S}$ & The delayed insulin action in the liver & -& -\\
        $S_{sto}(t)$ & $\mathcal{S}$ &  The amount of solid carbohydrates in stomach & mg & -\\
        $Q_{sto}(t)$ & $\mathcal{S}$ & The amount of liquid carbohydrates in stomach & mg & -\\
        $Q_{gut}(t)$ & $\mathcal{S}$ & The amount of liquid carbohydrates in gut & mg & -\\
        $Ra(t)$ & $\mathcal{S}$ & The rate of glucose absorption in the blood & -& -\\
        $E(t)$ & $\mathcal{S}$ & The renal excretion of glucose & mg/dL & -\\
        $EGP(t)$ & $\mathcal{S}$ & The endogenous glucose production (EGP) & U/day & -\\
        $U_{id}(t)$ & $\mathcal{S}$ & The insulin-dependent utilization takes place in the remote compartment & -& -\\
        $CHO(t)$ & $\mathcal{S}$ & The amount of ingested carbohydrates & g & (0, 200)\\
        $a(t)$ & $\mathcal{A}$ &The insulin concentration of the insulin pump & U/h & (0, 30) \\
        
    \hline
    \end{tabular}
\end{table*}

\textbf{Parameters Description}: The parameters for the \textit{SimGlucoseEnv} ODE system are derived from human physiological data, accommodating the creation of 100 virtual adults, 100 adolescents, and 100 children profiles. Given the individual variability among patients, specific default values for these parameters are not assigned. Instead, the parameters vary across the simulated individuals to better reflect the diversity found in real-world populations. An adolescent virtual patient variable description is provided in Table \ref{table:glucose_parameters}. The environment includes 10 adolescents, 10 adults, and 10 children, each with a unique set of dynamic parameters. Full details can be found at \cite{man2014uva}.

\textbf{Reward Function}: The reward function calculates a numerical reward based on the current and subsequent basal glucose ($G_p$) levels in plasma and the status of the episode. The reward function is defined as $r(t) = r_{\text{risk}}(t) + r_{\delta}(t) + r_{o}(t)$. Reward components are described as follows.

 $r_{\text{risk}}$ is the risk indices \cite{fabris2016risk} with $\log_{10}$ transformation. This reward encourages the agent to take action to reduce diabetes-related risks.
\begin{equation}
r_{\text{risk}}(t) = -\log_{10}\left( \left[1.509 \left( \ln(G_p(t))^{1.084} - 5.381 \right) \right]^2 \right) 
\end{equation}
    
$r_{\Delta}$ is the fluctuation reward which penalises abrupt fluctuations in basal glucose levels. This reward discourages actions that lead to significant variations in glucose levels, promoting a more stable glucose management.
\begin{equation}
r_{\Delta}(t) = 
\begin{cases} 
0 & \text{if } \Delta G_p(t) < 30, \\
-\frac{1}{30} (\Delta G_p(t) - 30), & \text{if } 30 \leq \Delta G_p(t) < 60, \\
-1 & \text{if } \Delta G_p(t) \geq 60,
\end{cases}
\end{equation}
where \( \Delta G_p(t) = G_p(t) - G_p(t-1) \).

$r_o$ is the reward for the outcome, which is given to encourage the agent to avoid early termination due to hypoglycemia or extreme hyperglycemia (i.e., when $G_p(t) < 10$ or $G_p(t) > 600$).
    
    \begin{equation}
    r_o(t) = 
    \begin{cases} 
    100 & \text{if $\forall G_p(1:T) \in [10, 600]$}, \\
    -100 & \text{if $\exists G_p(t) < 10$ or $\exists G_p(t) > 600$,} \\
    0 & \text{otherwise.}
    \end{cases}
    \end{equation}

\section{Results}
In this section, we first highlight the key features of the proposed DTR-Bench platform. DTR-Bench aims to enable more realistic simulations for evaluating Dynamic Treatment Regimes. Next, we present empirical results from the testing of various reinforcement learning algorithms in the four simulation environments included in DTR-Bench. These results demonstrate the improvements in realism and evaluation capabilities gained by the enhancements we have introduced in DTR-Bench. The findings illustrate how DTR-Bench allows for more robust testing of RL-driven personalised medicine approaches in silico, helping to address the challenges of transitioning these algorithms from simulation to real clinical settings.

\subsection{Feature Highlights for \textit{DTR-Bench} Platform}
DTR-Bench is an open-source repository designed for the development and evaluation of standardised RL-DTRs. This platform offers several notable features that make it a valuable tool.

\begin{itemize}
    \item \textbf{Evaluation Framework for RL Algorithms:} \textit{DTR-Bench} serves as an advanced evaluation platform, designed to accurately simulate complex healthcare scenarios. It incorporates critical factors such as noise, PK/PD variance, hidden variables, and missing data to enhance simulation realism.
    \item \textbf{Standard API Integration:} The platform is integrated with the standard \textit{Gym}\cite{1606.01540}, \textit{Gymnasium}\cite{towers_gymnasium_2023} and \textit{tianshou}\cite{tianshou} package API, ensuring ease of use and compatibility with the wider RL research and development ecosystem. This facilitates a user-friendly, plug-and-play approach for the customised development of RL policies.
    \item \textbf{Customised Development of RL Policies:} \textit{DTR-Bench} supports plug-and-play functionality for tailored algorithm design and experimentation, empowering researchers to innovate with new algorithms or modify existing ones.
    \item \textbf{Automatic Hyperparameter optimisation:} The platform includes features for convenient auto-hyperparameter optimisation with the integration of \textit{Optuna}\cite{akiba2019optuna}, streamlining the model tuning process for optimal performance.
    \item \textbf{Visualisation Support:} \textit{DTR-Bench} provides visualisation tools that allow users to visually assess the performance of RL algorithms and decision-making processes within simulation environments.
\end{itemize}

\subsection{Benchmark Results}
We evaluate discrete-action algorithms (DQN, DDQN, DDQN-duel, C51, dSAC), their sequential variants (rDQN, rDDQN, rDDQN-duel, rC51, discrete-rSAC), and continuous-action algorithms (DDPG, TD3, SAC) on  \textit{AhnChemoEnv}, \textit{GhaffariCancerEnv}, \textit{OberstSepsisEnv}, and \textit{SimGlucoseEnv} with 4 settings, shown in Table \ref{tab:ahn results}, \ref{tab:gha results}, \ref{tab:sepsis results}, and, \ref{tab:glucose results}, respectively. All results are based on optimised hyperparameters. Testing results are provided with `mean $\pm$ std' from 25,000 testing episodes on five seeds.

\begin{table*}[!t]
\centering
\caption{\textbf{Reward performance of various RL algorithms on four simulation environments under realist DTR enhancements ($\hat{p}$, $\hat{p_1}$, $\hat{p_2}$, $\hat{p_3}$).} Numbers, the higher, the better. The best reward for each column is highlighted in red, and the second best is highlighted in blue. $\pi_{b} $ refers to the highest reward among the 'baseline policies', i.e., random, zero-drug, and max-drug policies. $\hat{p}$, $\hat{p_1}$, $\hat{p_2}$, $\hat{p_3}$ denote the original, PK/PD variant, PK/PD+noise and PK/PD+noise+missing value enhancement, respectively. More details of the introduction of  $\hat{p_1}$, $\hat{p_2}$, $\hat{p_3}$ is provided in the Method section.}
\label{tab:overall}
\begin{subtable}{0.9\textwidth}
    \centering
    \small
    \caption{\textit{AhnChemoEnv}.}
    \label{tab:ahn results}
\begin{tabular}{l|c c c c} 
    \hline
    Policy & $\hat{p}$ & $\hat{p_1}$ & $\hat{p_2}$ & $\hat{p_3}$ \\
    \hline
$\pi_b$ & $62.29 \pm 12.62$ & $64.13 \pm 319.24$ & $60.60 \pm 324.14$ & $60.60 \pm 324.14$ \\
DQN & $214.18 \pm 141.64$ & $129.24 \pm 367.54$ & $95.76 \pm 389.00$ & $126.55 \pm 394.22$ \\
DDQN & \textcolor{red}{$370.61 \pm 18.36$} & \textcolor{red}{$170.02 \pm 375.16$} & \textcolor{red}{$134.36 \pm 376.11$} & \textcolor{blue}{$131.96 \pm 389.68$} \\
DDQN-duel & $248.59 \pm 66.77$ & $95.04 \pm 349.67$ & $105.22 \pm 344.06$ & $87.44 \pm 354.02$ \\
C51 & $144.30 \pm 90.10$ & $108.05 \pm 344.97$ & $122.53 \pm 377.36$ & $128.05 \pm 386.88$ \\
dSAC & $114.06 \pm 82.68$ & \textcolor{blue}{$137.88 \pm 404.36$} & $124.14 \pm 385.20$ & \textcolor{red}{$132.34 \pm 357.61$} \\
rDQN & $267.78 \pm 62.65$ & $113.90 \pm 352.34$ & $120.34 \pm 367.73$ & $91.13 \pm 368.63$ \\
rDDQN & $317.34 \pm 35.24$ & $132.60 \pm 350.05$ & $101.11 \pm 372.60$ & $88.92 \pm 342.95$ \\
rC51 & $242.18 \pm 82.29$ & $36.58 \pm 277.86$ & $64.17 \pm 331.92$ & $73.49 \pm 321.97$ \\
drSAC & \textcolor{blue}{$341.01 \pm 24.63$} & $126.53 \pm 378.35$ & $99.46 \pm 391.72$ & $102.30 \pm 389.97$ \\
DDPG & $113.28 \pm 55.93$ & $80.95 \pm 327.33$ & \textcolor{blue}{$124.67 \pm 376.75$} & $102.50 \pm 349.72$ \\
TD3 & $90.32 \pm 54.04$ & $82.10 \pm 339.29$ & $87.38 \pm 344.40$ & $76.97 \pm 347.48$ \\
SAC & $151.72 \pm 88.35$ & $89.09 \pm 338.32$ & $115.99 \pm 363.45$ & $108.80 \pm 374.35$ \\
\hline
\end{tabular}
\end{subtable}
\hfill
\begin{subtable}{0.9\textwidth}
    \centering
    \small
    \caption{\textit{GhaffariCancerEnv}}
    \label{tab:gha results}
\begin{tabular}{l|c c c c} 
    \hline
    Policy & $\hat{p}$ & $\hat{p_1}$ & $\hat{p_2}$ & $\hat{p_3}$\\
    \hline
$\pi_b$ & $-90.78 \pm 2.96$ & $-90.79 \pm 2.67$ & $-90.76 \pm 2.72$ & $-90.76 \pm 2.72$ \\
DQN & $134.24 \pm 26.45$ & $128.41 \pm 20.04$ & $131.14 \pm 28.43$ & $90.41 \pm 54.10$ \\
DDQN & $86.94 \pm 89.85$ & $132.76 \pm 34.16$ & $126.40 \pm 18.41$ & $115.26 \pm 32.23$ \\
DDQN-duel & $129.26 \pm 69.11$ & $130.71 \pm 25.39$ & $125.29 \pm 17.78$ & $113.70 \pm 26.45$ \\
C51 & \textcolor{red}{$160.91 \pm 62.61$} & \textcolor{red}{$136.28 \pm 34.87$} & \textcolor{red}{$139.89 \pm 34.00$} & \textcolor{blue}{$121.88 \pm 58.38$} \\
dSAC & $93.65 \pm 73.98$ & $127.65 \pm 12.29$ & $75.23 \pm 81.40$ & $96.86 \pm 70.69$ \\
rDQN & $63.66 \pm 85.66$ & $126.53 \pm 13.36$ & $124.89 \pm 16.31$ & $104.80 \pm 55.21$ \\
rDDQN & $57.40 \pm 87.30$ & $125.09 \pm 20.41$ & $108.64 \pm 43.99$ & $67.04 \pm 72.38$ \\
rC51 & $-82.88 \pm 16.37$ & $-74.74 \pm 22.22$ & $-79.81 \pm 19.60$ & $-83.36 \pm 16.91$ \\
drSAC & \textcolor{blue}{$143.45 \pm 18.09$} & \textcolor{blue}{$134.60 \pm 16.05$} & \textcolor{blue}{$139.64 \pm 15.34$} & $114.72 \pm 36.64$ \\
DDPG & $-33.52 \pm 84.31$ & $-70.15 \pm 39.87$ & $-29.19 \pm 73.69$ & $-48.61 \pm 76.66$ \\
TD3 & $62.81 \pm 76.83$ & $78.30 \pm 86.14$ & $90.96 \pm 75.44$ & $107.61 \pm 36.77$ \\
SAC & $127.06 \pm 4.53$ & $127.26 \pm 6.79$ & $124.99 \pm 8.52$ & \textcolor{red}{$124.73 \pm 24.80$} \\
    \hline
\end{tabular}
\end{subtable}
\hfill
\begin{subtable}{0.9\textwidth}
    \centering
    \small
    \caption{\textit{OberstSepsisEnv}}
    \label{tab:sepsis results}
\begin{tabular}{l|c c c c} 
    \hline
    Policy & $\hat{p}$ & $\hat{p_1}$ & $\hat{p_2}$ & $\hat{p_3}$\\
    \hline
$\pi_b$ & $-0.81 \pm 0.58$ & $-0.82 \pm 0.57$ & $-0.82 \pm 0.58$ & $-0.82 \pm 0.58$ \\
DQN & $0.15 \pm 0.73$ & $0.12 \pm 0.74$ & $0.08 \pm 0.71$ & $-0.08 \pm 0.46$ \\
DDQN & \textcolor{blue}{$0.17 \pm 0.73$} & $0.15 \pm 0.73$ & \textcolor{blue}{$0.09 \pm 0.73$} & $-0.12 \pm 0.57$ \\
DDQN-duel & $0.16 \pm 0.75$ & \textcolor{red}{$0.16 \pm 0.72$} & \textcolor{red}{$0.09 \pm 0.72$} & $-0.08 \pm 0.52$ \\
C51 & $0.13 \pm 0.72$ & $0.08 \pm 0.64$ & $0.05 \pm 0.65$ & \textcolor{red}{$-0.05 \pm 0.27$} \\
dSAC & $0.16 \pm 0.71$ & \textcolor{blue}{$0.16 \pm 0.72$} & $0.06 \pm 0.65$ & \textcolor{blue}{$-0.06 \pm 0.31$} \\
rDQN & $-0.11 \pm 0.77$ & $-0.22 \pm 0.62$ & $-0.19 \pm 0.70$ & $-0.27 \pm 0.53$ \\
rDDQN & $-0.17 \pm 0.74$ & $-0.15 \pm 0.77$ & $-0.17 \pm 0.74$ & $-0.29 \pm 0.53$ \\
rC51 & $-0.68 \pm 0.51$ & $-0.35 \pm 0.48$ & $-0.16 \pm 0.37$ & $-0.43 \pm 0.53$ \\
drSAC & \textcolor{red}{$0.17 \pm 0.52$} & $-0.04 \pm 0.26$ & $-0.08 \pm 0.30$ & $-0.15 \pm 0.36$ \\
    \hline
\end{tabular}
\end{subtable}
\hfill
\begin{subtable}{0.9\textwidth}
    \centering
    \small
    \caption{\textit{SimGlucoseEnv}}
    \label{tab:glucose results}
\begin{tabular}{l|c c c c} 
    \hline
    Policy & $\hat{p}$ & $\hat{p_1}$ & $\hat{p_2}$ & $\hat{p_3}$\\
    \hline
$\pi_b$ & $-187.86 \pm 0.00$ & $-182.86 \pm 45.51$ & $-184.36 \pm 50.76$ & $-184.36 \pm 50.76$ \\
DQN & $119.08 \pm 34.50$ & \textcolor{red}{$-144.56 \pm 96.74$} & $-160.46 \pm 102.33$ & \textcolor{red}{$-150.80 \pm 102.16$} \\
DDQN & $105.29 \pm 12.21$ & $-151.88 \pm 95.48$ & $-156.85 \pm 102.14$ & \textcolor{blue}{$-152.64 \pm 108.70$} \\
DDQN-duel & \textcolor{blue}{$123.52 \pm 55.72$} & $-157.01 \pm 95.46$ & $-189.87 \pm 105.58$ & $-177.73 \pm 107.16$ \\
C51 & $109.81 \pm 44.57$ & $-153.01 \pm 86.86$ & $-177.68 \pm 86.77$ & $-171.52 \pm 97.87$ \\
dSAC & $-16.28 \pm 107.64$ & $-201.11 \pm 89.80$ & $-179.21 \pm 98.02$ & $-168.73 \pm 102.64$ \\
rDQN & $-7.26 \pm 90.57$ & $-168.98 \pm 91.97$ & $-167.74 \pm 102.03$ & $-167.60 \pm 89.42$ \\
rDDQN & $114.78 \pm 55.44$ & $-163.81 \pm 89.98$ & $-156.11 \pm 96.02$ & $-156.61 \pm 96.38$ \\
rC51 & $-149.60 \pm 164.17$ & $-195.49 \pm 92.99$ & $-165.10 \pm 98.97$ & $-167.35 \pm 89.13$ \\
drSAC & $119.10 \pm 60.38$ & \textcolor{blue}{$-147.45 \pm 97.24$} & \textcolor{red}{$-151.61 \pm 95.20$} & $-160.24 \pm 96.11$ \\
DDPG & $-304.35 \pm 58.25$ & $-233.84 \pm 77.66$ & $-248.16 \pm 74.88$ & $-248.52 \pm 74.94$ \\
TD3 & $-244.18 \pm 120.52$ & $-233.58 \pm 77.77$ & $-248.63 \pm 75.43$ & $-234.00 \pm 82.20$ \\
SAC & \textcolor{red}{$170.60 \pm 1.49$} & $-151.57 \pm 94.41$ & \textcolor{blue}{$-155.14 \pm 106.24$} & $-152.98 \pm 105.25$ \\
    \hline
\end{tabular}
\end{subtable}
\end{table*}

\subsubsection{Performance across environments}
We've noted a consistent decrease in algorithm performance when accounting for PK/PD variance, noise, and missing values. This highlights the critical role of these factors in the thorough evaluation of the robustness of RL algorithms. This declining trend underscores the importance of incorporating real-world challenges, such as dynamic treatment regimes (DTR), to accurately gauge the resilience and adaptability of RL methods in practical scenarios. Notably, when considering the \textit{SimGlucoseEnv}, no algorithm achieved a positive reward under PK/PD variance, revealing the significant impact of individualised PK/PD models across different demographics of patients (elderly, adult, children). This variability underscores the specific challenges of implementing RL in DTR and emphasises the need for more personalised algorithmic approaches.

\subsubsection{Comparing discrete, continuous and sequential algorithms}
Our analysis highlights a varied performance landscape in which no single algorithm consistently outshines others in all environments and settings. In particular, discrete-action algorithms like C51 and dSAC demonstrate strong performance, particularly evident in the \textit{GhaffariCancerEnv} and \textit{SimGlucoseEnv}, suggesting their adaptability to the specific dynamics of these environments. However, the performance of their RNN variants does not uniformly surpass their MLP-based counterparts, indicating that incorporating medical history through sequential modelling doesn't always lead to improved predictive or decision-making accuracy. This discrepancy is notable in environments such as \textit{OberstSepsisEnv}, where dynamics rely less on temporal patterns.

\subsubsection{Algorithm Performance across settings}

\textbf{Medical History is not always helpful}: The expectation that incorporating sequential medical history through RNN-based models would universally enhance performance was unmet. Particularly in environments such as \textit{OberstSepsisEnv}, where the causal structure does not heavily depend on temporal dynamics, these models often faltered, emphasising the need for environment-specific model selection.

\textbf{Generalisation to PK/PD Variance:} Algorithms like DDQN-duel and drSAC showed notable adaptability, particularly in \textit{GhaffariCancerEnv} and \textit{AhnChemoEnv}. For instance, DDQN-duel's performance in Setting 2 across these environments was strong, indicating its robustness to PK/PD variance. Similarly, drSAC maintained commendable performance across multiple settings, highlighting its potential in handling varied treatment effects and patient responses, which is crucial for healthcare applications. However, certain algorithms struggled significantly with PK/PD variance. For example, rDQN and rC51 exhibited marked performance declines in \textit{SimGlucoseEnv} and \textit{OberstSepsisEnv}, indicating challenges in adapting to PK/PD variances within these environments. The performance of algorithms such as DQN and DDQN also saw considerable drops when transitioning from Setting 1 (original) to Setting 2 (with PK/PD variance), underscoring the difficulty in generalising these variances without specific adaptations.

\textbf{Robustness Against Noise:} C51's resilience, particularly in handling large noise and missing observations in \textit{OberstSepsisEnv} and \textit{GhaffariCancerEnv}, exemplifies its robustness. Its consistent performance in Setting 5 across various environments showcases its reliability amidst significant data challenges. Conversely, algorithms such as rDQN and rDDQN demonstrated considerable vulnerability to noise, with their performance notably deteriorating in noisy settings. This is evident in environments like \textit{SimGlucoseEnv}, where the addition of noise led to substantial performance drops. These observations highlight certain algorithms' challenges in maintaining efficacy under noisy conditions, emphasising the need for robust design and tuning to counteract environmental noise.

\textbf{Handling Missing Values:} The challenge of missing observations significantly affected all environments, reflecting complexities similar to real-world clinical data. The comparison between Setting 4 and Setting 5 reveals interesting dynamics. Although most models experienced performance degradation, the impact varied between algorithms. C51 and dSAC showed relative resilience, with less pronounced performance drops between these settings. This suggests these models' potential to deal with incomplete data, a crucial trait for healthcare applications. However, algorithms like rDQN and rDDQN saw more significant performance decreases, indicating their sensitivity to missing data. Interestingly, some models displayed a slight improvement or lesser performance drop in Setting 5 compared to Setting 4, such as drSAC in \textit{GhaffariCancerEnv} and SAC in \textit{SimGlucoseEnv}. However, these improvements are insignificant because of the high variance. This analysis underlines the importance of selecting and tailoring algorithms that are inherently more capable of handling the specific challenges presented by missing observations in clinical datasets.

\section{Discussion}
Our research delves into the performance of RL algorithms within various DTR settings, highlighting a notable performance decline when faced with real-world challenges such as PK/PD variance, noise, and missing data. This highlights the critical need to incorporate these practical challenges into evaluating RL algorithms better to reflect their potential effectiveness in real-world healthcare applications. 

Through our investigation, we discovered several intriguing findings. Contrary to expectations, RNN-based models did not consistently outperform their MLP counterparts. This outcome challenges the assumption that RNNs, with their capacity to process sequential data, would inherently provide superior performance in healthcare contexts. Furthermore, some algorithms, particularly C51 and drSAC, demonstrated remarkable resilience against volatility introduced by PK/PD variance and noise. These observations suggest that certain algorithmic features may be better suited to managing the complexities inherent in healthcare data.

The study also acknowledges several limitations that must be considered when interpreting the results. As the initial open-source platform and benchmark for RL-based DTR, \textit{DTR-Bench} only collected four environments, potentially limiting the external validity of the diseases we have yet to cover. Furthermore, our focus was limited to off-policy RL algorithms, and we did not investigate the essential aspect of safe exploration within treatment regimes.

Looking ahead, there is a clear imperative for further research into more advanced RL models equipped to navigate the unpredictable and varied landscape of healthcare data. Moreover, the creation and utilization of sophisticated medical simulations that more closely resemble real-world conditions would significantly contribute to the validation and refinement of these models. Such advancements are crucial for enhancing the credibility and applicability of research outcomes in this domain.


\section{Conclusion}

This paper introduced \textit{DTR-Bench}, an \textit{in silico} benchmark platform aimed at addressing the critical challenges associated with the development of DTR through RL in healthcare settings. \textit{DTR-Bench} is designed to provide a realistic testing ground for RL algorithms by incorporating key clinical factors such as PK/PD variance, noise, and the complexities of missing data. This approach allows for a more rigorous evaluation of the capabilities and limitations of RL within various medical treatments, including cancer chemotherapy, radiotherapy, glucose management in diabetes, and sepsis treatment. Our analysis of several state-of-the-art RL algorithms in these diverse simulation environments has highlighted significant variations in performance, underscoring the importance of customised algorithmic development for healthcare applications. 



\bibliographystyle{IEEEtran}
\bibliography{refs}

\newpage
\input{appendix}

\end{document}

%% file: appendix.tex
\appendix


\begin{table*}[h]
    \small
    \centering
        \caption{\textbf{Hyperparameter search space.} 'Common' hyperparameters are model-agnostic and used identically in all runs. Stack number means the number of stacking frames for RNN policy. In non-RNN case, the stack number is 1. }
    \label{tab:hyperparameters}
    \begin{tabular}{c l c } \hline 
         Type&  Name & Space\\ 
         \hline
         \multirow{14}{*}{Common} & Learning Rate & $[10^{-3}, 10^{-4},$ $10^{-5}, 10^{-6}]$\\ 
          & Batch Size &  [128, 256, 512, 1024] \\
          & Stack Number & [20, 50, 100] \\
          & Batch Normalisation & [True, False]\\
          & Dropout & [0.0, 0.25, 0.5] \\
          & Target Update Frequency & [1, 1000, 5000]\\
          & Update Per Step & [0.1, 0.5] \\
          & Update Actor Frequency & [1, 5, 10] \\
          & Step Per Collect & [50, 100] \\
          & Exploration Noise & [0.1, 0.2, 0.5] \\
          & Estimation Step & 1 \\
          & $\gamma$ & 0.95 \\
          & $\tau$ & 0.001 \\
          \hline
         \multirow{3}{*}{DQN Series} & $\epsilon_\text{test}$ & 0.005 \\
          & $\epsilon_{\text{train},0}$ & 1 \\
          & $\epsilon_{\text{train},-1}$ & 0.005 \\
          \hline
         \multirow{6}{*}{C51, rC51} & $v_\text{min}$ & [-20, -10, -5] \\
          & $v_\text{max}$ & [5, 10, 20] \\
          & Number of Atoms & 51 \\
          & $\epsilon_\text{test}$ & 0.005 \\
          & $\epsilon_{\text{train},0}$ & 1 \\
          & $\epsilon_{\text{train},-1}$ & 0.005 \\
         \hline
         SAC, rSAC, drSAC & $\alpha$ & [0.05, 0.1, 0.2] \\
         \hline
         \multirow{3}{*}{REDQ} & $\alpha$ & [0.05, 0.2, 0.4] \\
          & Ensemble Size & 10 \\
          & Actor Decay & 20 \\
         \hline
         \multirow{2}{*}{TD3} & Noise Clip & 0.5 \\
         & Policy Noise & 0.2 \\
        \hline
    \end{tabular}

\end{table*}

\input{tables/table_parameter_ahnchemo}

\input{tables/table_parameter_ghaffaricancer}

\input{tables/table_parameter_simglucose}

%% file: tables/table_parameter_ahnchemo.tex
\begin{table*}[h]
\small
\centering
\caption{Parameters of the \textit{AhnChemoEnv} ODEs system}
\label{table:ahn_parameters}
\begin{tabular}{c c p{8cm} c}
\hline
\textbf{Parameter} & \textbf{Unit} & \textbf{Description} & \textbf{Default Value} \\
\hline
$a_i$ & - & Fraction cell kill (drug effectiveness) & $a_1=0.2$, $ a_2 = 0.3$,  $ a_3 = 0.1$ \\
$b_i$ & - & Carrying capacities (inverse) & $b_1=1.0$, $ b_2 = 1.0$ \\
$c_i$ & - & Competition terms between cells & $c_1=1.0$, $c_2=0.5$, $c_3=1.0$,  $c_4=1.0$  \\
$d_1, d_2$ & - & Death rates of immune cells and drug effect & $d_1=0.2$, $d_2 = 1.0$ \\
$r_1, r_2$ & - & Per unit growth rates of tumour and normal cells & $r_1 =1.5$, $r_2 = 1.0$ \\
$s$ & - & Immune source rate in the absence of a tumour & $s=0.33$ \\
$\rho$ & - & Immune response rate & $\rho=0.01$\\
\hline
\end{tabular}
\end{table*}

%% file: tables/table_parameter_ghaffaricancer.tex
\begin{table*}
\centering
\small
\caption{Parameters of the \textit{GhaffariCancerEnv} ODEs system, part I} \label{table:gha params}
\begin{tabular}{c c p{8cm} c}
\hline
\textbf{Parameter} & \textbf{Units} & \textbf{Description} & \textbf{Default value} \\
\hline
$a_{1}$ & day$^{-1}$ & Primary tumour growth rate & $4.31 \times 10^{-1}$ \\
$b_{1}$ & cells$^{-1}$ & $1 / b$ is tumour carrying capacity (in primary tumour) & $1.02 \times 10^{-9}$ \\
$c_{1}$ & cell$^{-1}$day$^{-1}$ & Fractional (non)-ligand-transduced tumour cell kill by NK cells (in primary tumour) & $6.41 \times 10^{-11}$ \\
$d_{1}$ & day$^{-1}$ & Saturation level of fractional tumour cell kill by CD8$^+$ T cells. Primed with ligand-transduced cells, challenged with ligand-transduced cells (in primary tumour) & 2.34 \\
$I$ & - & Exponent of fractional tumour cell kill by CD8$^{+}$ T cells. Primed with ligand-transduced cells, challenged with ligand-transduced cells (in primary tumour) & 2.09 \\
$S$ & - & Steepness coefficient of the tumour-(CD8$^{+}$ T cell) lysis term D. Primed with ligand-transduced cells, challenged with ligand-transduced cells (in primary tumour) & $8.39 \times 10^{-2}$ \\
$e_{1}$ & day$^{-1}$ & Fraction of circulating lymphocytes that become NK cells (in primary tumour) & $2.08 \times 10^{-1}$ \\
$f_{1}$ & day$^{-1}$ & Death rate of NK cells (in primary tumour) & $4.12 \times 10^{-3}$ \\
$p_{1}$ & cell$^{-1}$day$^{-1}$ & NK cell inactivation rate by tumour cells (in primary tumour) & $3.42 \times 10^{-4}$ \\
$m_{1}$ & day$^{-1}$ & Death rate of CD8$^{+}$ T cells (in primary tumour) & $2.04 \times 10^{-2}$ \\
$j_{1}$ & day$^{-1}$ & Maximum CD8$^{+}$ T cell recruitment rate. Primed with ligand-transduced cells, challenged with ligand-transduced cells (in primary tumour) & $2.49 \times 10^{-2}$ \\
$k_{1}$ & cells$^{2}$ & Steepness coefficient of the CD8$^{+}$ T cell recruitment curve (in primary tumour) & $3.66 \times 10^{7}$ \\
$q_{1}$ & cell$^{-1}$day$^{-1}$ & CD8$^{+}$ T cell inactivation rate by tumour cells (in primary tumour) & $1.42 \times 10^{-4}$ \\
$r_{11}$ & cell$^{-1}$day$^{-1}$ & Rate at which CD8$^{+}$ T cells are stimulated to be produced as a result of tumour cells killed by NK cells (in primary tumour) & $1.1 \times 10^{-7}$ \\
$r_{12}$ & cell$^{-1}$day$^{-1}$ & Rate at which CD8$^{+}$ T cells are stimulated to be produced as a result of tumour cells interacting with circulating lymphocytes (in primary tumour) & $6.5 \times 10^{-11}$ \\
$u_{1}$ & cell$^{-2}$day$^{-1}$ & Regulatory function by NK cells of CD8$^{+}$ T-cells (in primary tumour) & $3 \times 10^{-10}$ \\
$K_{1 T}$ & day$^{-1}$ & Fractional tumour cell kill by chemotherapy (in primary tumour) & 100 \\
$K_{1 L}$, $K_{1 N}$ & day$^{-1}$ & Fractional immune cell (in primary tumour) & 10 \\
$K_{1C}$ & day$^{-1}$ & Fractional immune cell kill by chemotherapy (in primary tumour) & 10 \\
$\alpha$ & cellday$^{-1}$ & Constant source of circulating lymphocytes & $7.5 \times 10^{8}$  \\
$\beta$ & day$^{-1}$ & Natural death and differentiation rate of circulating lymphocytes & $1.2 \times 10^{2}$ \\
$\mu$ & day$^{-1}$ & Rate of chemotherapy drug decay & $9 \times 10^{-1}$ \\
$a_{2}$ & day$^{-1}$ & Secondary tumour growth rate & 5  \\
$b_{2}$ & cells$^{-1}$ & $1 / b$ is tumour carrying capacity (in secondary tumour) & $1 \times 10^{-7}$  \\
$c_{2}$ & cell$^{-1}$day$^{-1}$ & Fractional (non)-ligand-transduced tumour cell kill by NK cells (in secondary tumour) & $6.41 \times 10^{-12}$ \\
$d_{2}$ & day$^{-1}$ & Saturation level of fractional tumour cell kills by CD8$^{+}$ T cells, primed with ligand-transduced cells, challenged with ligand-transduced cells (in secondary tumour) & 5 \\
$e_{2}$ & day$^{-1}$ & Fraction of circulating lymphocytes that become NK cells (in secondary tumour) & $2.08 \times 10^{-1}$ \\
$f_{2}$ & day$^{-1}$ & Death rate of NK cells (in secondary tumour) & $3.5 \times 10^{-2}$ \\
$p_{2}$ & cell$^{-1}$day$^{-1}$ & NK cell inactivation rate by tumour cells (in secondary tumour) & $1 \times 10^{-1}$ \\
$m_{2}$ & day$^{-1}$ & Death rate of CD8$^{+}$ T cells (in secondary tumour) & $1.8 \times 10^{-1}$ \\
$j_{2}$ & day$^{-1}$ & Maximum CD8$^{+}$ T cell recruitment rate, primed with ligand-transduced cells, challenged with ligand-transduced cells (in secondary tumour) & $1.6 \times 10^{-2}$ \\
\hline
\end{tabular}

\end{table*}

\begin{table*}
\centering
\small
\caption{Parameters of the \textit{GhaffariCancerEnv} ODEs system, part II} \label{table:gha params2}
\begin{tabular}{c c p{8cm} c}
\hline
\textbf{Parameter} & \textbf{Units} & \textbf{Description} & \textbf{Default value} \\
\hline
$k_{2}$ & cells$^{2}$ & Steepness coefficient of the CD8$^{+}$ T cell recruitment curve (in secondary tumour) & $3.66 \times 10^{7}$ \\
$q_{2}$ & cell$^{-1}$day$^{-1}$ & CD8$^{+}$ T cell inactivation rate by tumour cells (in secondary tumour) & $1 \times 10^{-1}$ \\ 
$r_{21}$ & cell$^{-1}$day$^{-1}$ & Rate at which CD8$^{+}$ T cells are stimulated to be produced as a result of tumour cells killed by NK cells (in secondary tumour) & $2 \times 10^{-1}$ \\
$r_{22}$ & cell$^{-1}$day$^{-1}$ & Rate at which CD8C T cells are stimulated to be produced as a result of interacting with circulating lymphocytes (in secondary tumour) & $7.5 \times 10^{11}$ \\
$u_{2}$ & cell$^{-2}$day$^{-1}$ & Regulatory function by NK cells of CD8$^{+}$ T-cells (in secondary tumour) & $3 \times 10^{-10}$ \\
$K_{2T}$ & day$^{-1}$ & Fractional tumour cell kill by chemotherapy (in secondary tumour) & 100 \\
$K_{2L}, K_{2N}$ & day$^{-1}$ & Fractional immune cell kill by chemotherapy (in secondary tumour) & 10 \\
$K_{2C}$ & day$^{-1}$ & Fractional immune cell kill by chemotherapy (in secondary tumour) & 10 \\
$\gamma_{1}$ & cell$^{-1}$day$^{-1}$ & Restoration and recovery rate of damaged cells irradiated & 0.04 \\
$\gamma_{2}$ & cell$^{-1}$day$^{-1}$ & Restoration and recovery rate of damaged cells irradiated & 0.1 \\
$\gamma_{3}$ & cell$^{-1}$day$^{-1}$ & Restoration and recovery rate of damaged cells irradiated & 0.1 \\
$\varepsilon$ & - & Part of radiation to damage healthy cells & 0.05 \\
$\alpha_{1}$ & - & Rate at which cancer leaves the primary site & 0.0001 \\
$\alpha_{2}$ & - & Rate at which cancer cells reach and interact at the secondary site after having left the primary site & 0.00001 \\
$W_{1T}$ & cells$^{2}$ & Determine the speeds at which cancer cells, in the absence of competition and predation, reach carrying capacity in primary & 0.01 \\
$w_{1N}, w_{1L}, w_{1L}$ & cells$^{2}$ & Determine the speeds at which immune cells, in the absence of site competition and predation, reach carrying capacity in primary & 1 \\
$W_{2T}$ & cells$^{2}$ & Determine the speeds at which cancer cells, in the absence of site competition and predation, reach carrying capacity in secondary & 1 \\
$w_{2N} w_{2L}$ & cells$^{2}$ & Determine the speeds at which immune cells, in the absence of site competition and predation, reach carrying capacity in secondary & 1\\
\hline
\end{tabular}

\end{table*}

%% file: tables/table_parameter_simglucose.tex
\begin{table*}[h]
\small
    \centering
    \caption{Parameters of \textit{SimGlucoseEnv} ODEs}
    \label{table:glucose_parameters}
    \begin{tabular}{c c p{8cm} c}
        \hline
        \textbf{Parameter} & \textbf{Unit} & \textbf{Description} & \textbf{Default Value} \\
        \hline
        \(k_{p1}\) & mg/kg/h & Rate parameter for endogenous glucose production & 11.5048 \\
        \(k_{p2}\) & 1/mg/dL & Sensitivity of glucose production to plasma glucose & 0.0233 \\
        \(k_{p3}\)  & 1/h & Sensitivity of glucose production to insulin action & 0.0233 \\
        \(k_{e1}\) & mg/kg/h & Parameter for renal glucose excretion & 0.0005 \\
        \(k_{e2}\) & mg/dL & Glucose renal excretion threshold & 339 \\
        \(V_{m0}\)  & mg/kg/h & Maximum glucose utilization rate & 5.9285 \\
        \(V_{mx}\)  & mg/kg/h & Sensitivity of glucose utilization to insulin action & 0.0747 \\
        \(K_{m0}\)  & mg/dL & Glucose concentration at half-maximum utilization rate & 260.8900 \\
        \(k_1\), \(k_2\) & 1/h & Transfer rate parameters for glucose between plasma and non-plasma compartment & \(k_1=0.0573\), \(k_2=0.0677\) \\
        \(p_{2u}\) & 1/h & Parameter for insulin action on glucose utilization & 0.0213 \\
        \(k_i\)  & 1/h & Parameter for insulin dynamics & 0.0089 \\
        \(k_{sto}\)  & 1/h & Stomach emptying rate constant for solid carbohydrates & 0.0159 \\
        \(k_{gut}\)  & 1/h & Transfer rate constant for carbohydrates from stomach to gut & 0.0159 \\
        \(k_{abs}\)  & 1/h & Absorption rate constant for carbohydrates from gut to plasma & 0.0910 \\
        \(f\) & - & Bioavailability factor for glucose & 0.9 \\
        \(BW\) & kg & Body weight, used to normalize certain rates & 68.7060 \\
        \hline
    \end{tabular}
\end{table*}